\documentclass[prodmode,acmtap]{acmlarge}

\acmVolume{1}
\acmNumber{1}
\acmArticle{1}
\articleSeq{1}
\acmYear{2016}
\acmMonth{1}

\usepackage{amsmath}
\usepackage{amssymb}
\usepackage{amsfonts}
\usepackage{graphicx}
\usepackage{amstext}
\usepackage{subfig}
\usepackage{siunitx}
\usepackage[ruled]{algorithm2e}
\SetAlFnt{\algofont}
\SetAlCapFnt{\algofont}
\SetAlCapNameFnt{\algofont}
\SetAlCapHSkip{0pt}
\IncMargin{-\parindent}
\renewcommand{\algorithmcfname}{ALGORITHM}

\title{Using Reinforcement Learning to Validate Empirical Game-Theoretic Analysis: A Continuous Double Auction Study}
\author{MASON WRIGHT \affil{University of Michigan}}

\begin{abstract}
Empirical game-theoretic analysis (EGTA) has recently been applied successfully to analyze the behavior of large numbers of competing traders in a continuous double auction market. Multiagent simulation methods like EGTA are useful for studying complex strategic environments like a stock market, where it is not feasible to solve analytically for the rational behavior of each agent. A weakness of simulation-based methods in strategic settings, however, is that it is typically impossible to prove that the strategy profile assigned to the simulated agents is stable, as in a Nash equilibrium. I propose using reinforcement learning to analyze the regret of supposed Nash-equilibrium strategy profiles found by EGTA. I have developed a new library of reinforcement learning tools, which I have integrated into an extended version of the market simulator from our prior work. I provide evidence for the effectiveness of our library methods, both on a suite of benchmark problems from the literature, and on non-equilibrium strategy profiles in our market environment. Finally, I use our new reinforcement learning tools to provide evidence that the equilibria found by EGTA in our recent continuous double auction study are likely to have only negligible regret, even with respect to an extended strategy space.
\end{abstract}


\acmformat{Mason Wright. 2016. Using Reinforcement Learning to Validate Empirical Game-Theoretic Analysis.}

\begin{document}

\maketitle

\section{Introduction} \label{intro}

Simulation-based modeling allows researchers to investigate the workings of complex multi-agent systems, where the aggregate effects of individual agents' behavior cannot be derived analytically. In strategic multi-agent settings like a stock market, researchers must not only develop a model of how agents interact with each other and their environment, but must also assign a rational policy to each agent. Empirical game-theoretic analysis (EGTA) provides a simulation-based method for finding strategy profiles that have negligible regret, with respect to a restricted set of strategies \cite{wellman2006methods}.

The continuous double auction (CDA) has long been a popular subject for game-theoretic research, due in part to its practical importance. Trillions of dollars are traded each year through continuous double auctions, and a commensurate effort has been expended in developing strategies for CDA trading \cite{schvartzman2009stronger}. EGTA has proven useful for studying what outcomes would likely arise in a CDA under hypothetical conditions, such as with or without a market maker agent in the market \cite{wah2015welfare}.

To study a counterfactual situation in a strategic setting, like what would happen if a CDA market were converted to a frequent call auction market, a researcher would use EGTA to search for stable strategy profiles (Nash equilibria) in each experimental condition. Next, the researcher could use conventional methods of empirical multi-agent simulation to compare outcomes when the different experimental conditions' equilibrium strategy profiles were enacted.

This general experimental design depends on the assumption that EGTA produces (approximately) strategically stable strategy profiles. If the profiles returned by EGTA were not equilibria, and in fact had high regret, then the outcomes of the experiment would be invalid, because the evidence would suggest that rational agents would not use those strategies or similar ones. EGTA is able to verify empirically whether a particular profile has positive regret, but only with respect to the limited set of strategies it has simulated. Because EGTA generates payoff data for only a small set of strategies, it leaves open the possibility that the profiles it returns are not Nash equilibria with respect to a wider strategy space. In fact, those profiles might have high regret.

In this study, I propose using reinforcement learning to search for strategies with higher payoff, against equilibrium profiles returned by EGTA. If learning methods can discover a strategy that yields significantly higher payoff, when played by one deviating agent against other agents playing some profile, this refutes the notion that the profile is a Nash equilibrium. In contrast, if a learning method has been shown to be capable of finding better-performing strategies against non-equilibrium profiles, and that method fails to find any strategy that outperforms a profile returned by EGTA; then this can be seen as evidence suggesting that the profile may be a Nash equilibrium, or at least have low regret. This evidence is an important supplement to EGTA because it goes beyond the small set of strategies for which EGTA generates sample payoffs, to the much larger policy space of a reinforcement learner.

In order to provide ample evidence for the strategic stability of a profile, in case it cannot be refuted by our learning algorithms, I have implemented several different reinforcement learners from prior literature. These include classical algorithms such as Q-learning and Sarsa, as well as the more complex POMCP algorithm, developed by Silver and Veness as a form of tree search for large partially observable Markov decision processes (POMDPs) \cite{silver2010monte}. I augment the classical algorithms with enhancements that have been shown to either accelerate or stabilize the learning process, including tile coding \cite{sherstov2005function}, experience replay \cite{lin1992self}, and eligibility traces \cite{loch1998using}. All of our implementations are thoroughly evaluated on benchmark environments from prior work, to validate their efficacy.

I present experimental results that support the claim that the Nash equilibria discovered in our market environment in a recent study \cite{wah2015welfare} are genuine equilibria, or at least have low regret. Granted, the empirical evidence of our learning agents cannot prove that there is no policy in the learners' policy space that would be a beneficial deviation from the EGTA profiles, because the learners' policy space is much too large to explore exhaustively. But I present two forms of experimental evidence in favor of the stability of the EGTA profiles. First, I show that our learning methods can consistently improve upon non-equilibrium profiles, and can improve upon EGTA profile performance when the learning agent is allowed to choose whether to buy or sell (as explained in Section \ref{environment}). Second, I show that our learning methods consistently reach a plateau in performance at or near the payoff of the EGTA profiles. This amounts to evidence strongly suggesting that the EGTA profiles have low regret, even in the larger policy space of the reinforcement learners.

\subsection{Contributions of the Author}

This report is given in advance of a preliminary exam, so it is appropriate to make clear what the author contributed, and what part of the work described here was done by others.

The market simulator used in this study was developed by Elaine Wah for use in multiple studies \cite{wah2013latency} \cite{wah2015welfare}. The continuous double auction market model used in this study is adopted directly from her work \cite{wah2015welfare}. The equilibria from empirical game theoretic analysis, which this paper seeks to validate, were found as part of that same project.

I extended the market simulator, to allow it to be used by a reinforcement learning algorithm, which required altering both the market scheduler and the component for keeping track of agents' payoffs. I also performed all of the reinforcement learning work presented in this paper, and I wrote the paper itself.

\subsection{Structure of the Paper}

The rest of the paper is organized as follows. Section \ref{related} presents related work on reinforcement learning and EGTA, and motivates our study of continuous double auctions. Section \ref{cdas} defines the continuous double auction as considered in our model. Section \ref{environment} provides a detailed description of the market model implemented by our simulator. Section \ref{strategies} explains the trading strategies used by agents in our model. Section \ref{egta} discusses empirical game theoretic analysis, including how it finds supposed equilibria, and why its results might not be true Nash equilibria. Section \ref{rl} presents the different reinforcement learning methods implemented and used in this work. Section \ref{experiments} describes the methods I used to validate our learning algorithm implementations and to test the stability of EGTA profiles. Section \ref{results} presents our results. Finally, Section \ref{discussion} wraps up the paper.

\section{Related Work} \label{related}

Prior work by Schvartzman and Wellman introduced the concept of interleaved reinforcement learning and EGTA \cite{schvartzman2009stronger}. The authors used Q-learning with tile coding to search for a continuous double auction trading strategy with greater mean payoff than a given strategy profile. The authors would then add the learned strategy to the strategy set in use by EGTA, and apply EGTA to find Nash equilibria in the expanded strategy set. This process would continue until Q-learning was no longer able to find a beneficial strategy deviation from the current EGTA equilibrium profile. The resulting profile is likely to be more strategically stable than a profile that is an equilibrium only with respect to the initial strategy set of EGTA. Schvartzman and Wellman later applied this technique to the Trading Agent Competition (TAC) Travel Game \cite{schvartzman2010learning}.

This work extends the methodology proposed by Schvartzman and Wellman, by using more sophisticated reinforcement learning techniques including POMCP, eligibility traces, and experience replay, as well as a greater variety of methods, such as off-policy (Q-learning), on-policy (Sarsa), and tree search based. Moreover, I apply reinforcement learning to a simulated CDA environment in which it may be more challenging for the learner to surpass the performance of the initial, supposed equilibrium profile found by EGTA. As a result, I employ RL methods with the goal of validating an equilibrium strategy profile from EGTA, as opposed to finding best responses to a non-equilibrium profile for inclusion in the next iteration of EGTA.

The CDA market model I use in this paper was introduced by Wah and Wellman in their study of latency arbitrage traders \cite{wah2013latency}. Wah and Wellman later extended the model with the addition of market maker agents, which supply liquidity to the market by offering to buy or sell shares at all times \cite{wah2015welfare}. The latter study, in particular, used strategic reasoning and EGTA to explore the economic effects of different experimental conditions (the presence or absence of a market maker). The authors used EGTA to find approximate Nash equilibrium strategy profiles for the background traders and market maker, and the results of the study were obtained by simulating the market with agents playing those equilibrium profiles. It would thus be useful to know whether a reinforcement learner, operating in a slightly expanded strategy space relative to the agents in the study, could find a strategy deviation that produces significantly higher expected payoff. If such a deviation is found, it refutes the equilibrium from EGTA, and casts doubt on the results of the study. If the learner reaches a plateau in performance at or near the payoff of the EGTA profile, then this is suggestive evidence that the EGTA profile has low regret, and the study's results are upheld.

The reinforcement learning problem addressed in this work creates a partially observable Markov decision process (POMDP) from a multi-agent game, by fixing the strategies of all other agents to a mixed strategy profile from EGTA. When all other agents play a particular strategy, the game becomes an optimization problem for the self agent. I draw insight from the literature on solving POMDPs for our learning approach. Silver and Veness proposed the POMCP method for solving a POMDP online through a generative model of the environment \cite{silver2010monte}, which allows us to use our market simulator as the generative model. As an alternative to building a tree representation of future states of the POMDP, I have also applied classical MDP solving methods like Q-learning and Sarsa$(\lambda)$ to our problem, treating the POMDP as if it were an MDP, with the observations as states. The eligibility traces of Sarsa$(\lambda)$ have been claimed to be useful in this approach \cite{loch1998using}. I have also used experience replay with Q-learning to accelerate convergence \cite{lin1992self}, a technique that has been popular lately for Q-learning combined with function approximation \cite{mnih2013playing}. I use tile coding in our Q-learning and Sarsa$(\lambda)$ implementations, to help the learner generalize across nearby observations in its discretized observation space, as suggested by Sherstov and Stone \cite{sherstov2005function}.

\section{Continuous Double Auctions} \label{cdas}

The continuous double auction (CDA) has long been the principal mechanism by which stocks and other securities are traded. Its use in such venues as the New York Stock Exchange has motivated considerable study of how rational agents should trade in the continuous double auction. Here I will present a formal description of the double auction as considered in this work, along with key parameters of auction state that afford agents information about how to trade.

The CDA is termed a double auction because it allows agents to participate as buyers or sellers, and agents trade with each other rather than with the house, as they would in a Dutch auction or ascending auction. The auction is continuous in time, as a transaction can occur any time a new buyer or seller arrives at the market and places an order. This distinguishes the CDA from sealed bid auctions, where there is a predetermined closing time when transactions, if any, must occur. Note that in our simulation study I will use discrete time steps, so that agents arrive at the market at integer-valued times, although with such fine granularity relative to the arrival frequencies that the market behaves similarly to a continuous-time market.

Now I define just how a CDA operates, as considered in this work. There exists a set of agents, or traders, each of which has an inventory $I$ comprising an integer-valued number of units of some good, say shares of stock. Any trader may use the CDA mechanism to buy or sell integer-valued sets of units. At any time from market open to close, a trader could arrive at the market and place an order.

Any order is labeled either BUY or SELL, is endowed with a positive integer-valued number of units to trade at maximum, and a positive integer-valued price at which to trade at worst. If the order matches a preexisting order (as explained below), a transaction occurs that fills all or part of the order, with the unfilled part placed in the \emph{limit order book}. The limit order book stores all BUY (SELL) orders, as a priority queue with the first items ordered primarily by highest (lowest) price, and secondarily by earliest time of being submitted.

If a new order to BUY (SELL) has a price no less than a SELL order in the book (no greater than a BUY order in the book), a transaction occurs at the price of the preexisting order in the book, filling the smaller order completely. Any remainder of the new order either transacts with the next-best matching order in the book, or is inserted into the book.

In our study, I consider only unit-size orders, meaning orders to buy or sell a single share. Thus, there is never any remainder left over after matching---an order either clears immediately with the best matching order, or it is added entirely to the order book.

Depending on a market's own special rules, traders may see various amounts of information about the state of a CDA in progress. For example, in a hypothetical stock market, some traders may be allowed to see the price and size of all orders in the limit order book, while others see only the price of the best bid and offer. In our simulation studies, I assume that each trader can sense only the price of the best bid and offer. Moreover, each trader is allowed to arrive at the simulated market only at certain times, drawn from an exponential distribution, as will be described in section \ref{environment}. A trader in our study knows how many time steps remain until the close of the auction, as well as its own current inventory. These and other state parameters known to the agent will be discussed in more detail in Section \ref{environment}.

\section{Market Environment} \label{environment}

In this work I consider a certain simulated stock market, with one stock being traded over a period of discrete time steps, from $0$ to some maximum $T$. I examine three distinct parameterized forms of this market, but in all cases there will be 25 \emph{background traders} with inherent demand for the stock encoded in private values, and 1 \emph{market maker} that has no private value, and seeks to profit by acting as a middleman in trades between other agents. Our market environment is drawn from a recent study of market maker effects on the outcomes for other traders, described at greater length in that work \cite{wah2015welfare}.

\subsection{Fundamental Value}

The market has a fundamental value $r_t$ for the stock at all times $t$, which is publicly known by the agents and evolves by a random walk. At the end of the simulation, each agent's inventory $I$ is automatically liquidated at the final fundamental value, $r_T$, yielding a marginal payoff of $I_T \times r_T$. Thus, the current fundamental value $r_t$ is useful as a noisy signal of the final payoff from holding a certain inventory.

The fundamental value evolves as a mean-reverting random walk, with random noise coming from the addition of a fixed-variance, zero-mean Gaussian term:
\begin{align*}
r_{t+1} \leftarrow \kappa \bar{r} + (1 - \kappa) \times (r_t + \mathcal{U}_t),
\end{align*}
where $\bar{r}$ is the long-run mean parameter, $\kappa \in [0, 1)$ is the mean reversion rate, and the shock term $\mathcal{U}$ is a zero-mean Gaussian with a  variance of $\sigma_s^2$. Note that the fundamental value is truncated at zero, so it will never be negative.

The agent's expectation of the final fundamental value, $\hat{r}_T$ hence evolves as another random walk \cite{wah2015welfare}. The expected value process is:
\begin{align*}
\hat{r}_T = \Big( 1 - (1 - \kappa)^{T - t}\Big) \times \bar{r} + (1 - \kappa)^{T - t} \times r_t.
\end{align*}

\subsection{Private Value}

Each background trader is given a vector of marginal private values at the beginning of the simulation. These values are constructed such that the more units in the trader's inventory $I$, the less the trader's private value will be for the next unit acquired. The values are drawn from a certain random distribution as described elsewhere \cite{wah2015welfare}. For this work, the exact nature of the distribution is less important than a few of its stylized properties. I note only that statistically, almost every background trader has a preferred final inventory in $\{-3, -2, -1, 0, 1, 2, 3\}$, and the most common preferred final inventory is $I = 0$ (i.e., the trader gets negative private value from gaining 1 share, and loses positive private value from giving up 1 share).

The market maker agent has no private value, so it has no inherent utility benefit for gaining or losing inventory. It seeks to make profits by buying shares for less than their expected final value (or selling for more).

All agents are assumed to be risk-neutral, so the goal is to maximize expected profit, without discounting large gains to mitigate risk.

\subsection{Arrival Process}

Our simulations proceed in discrete time steps, with \num[group-separator={,}]{1000} or \num[group-separator={,}]{4000} time steps per simulation run. Agents can potentially arrive at the market and place orders at any time, but each agent is assigned specific times to arrive by a random process. Specifically, each agent makes an independent, random draw from an exponential distribution to determine its next interarrival time. This time is then rounded and added to the agent's previous arrival time. Thus, the arrival process is implemented similarly to a geometric distribution over inter-arrival times. If the total is no greater than $T$, the simulation length, the agent will be scheduled to arrive at that time.

Each background trader arrives with interarrival times distributed as $\textrm{Exp}(\lambda_{BG})$, with expected interarrival time $\lambda_{BG}$. The market maker arrives at a different rate, with interarrival time distributed as $\textrm{Exp}(\lambda_{MM})$.

\subsection{Agent Senses and Actions}

When a background trader arrives, it can sense the number of times steps remaining $T - t$, the current BID and ASK prices (best prices in the limit order book to BUY or SELL), and whether it is assigned in this arrival to BUY or SELL. The agent computes the expected final fundamental value, $\hat{r}_T$, and the benefit in private value of gaining or giving up 1 unit, based on its current inventory $I$.

The agent's previous order, if any, is automatically canceled at this time. The agent can optionally submit a new order to BUY or SELL, depending on its assigned role, for 1 unit, for any positive price.

When a market maker arrives, it can sense all the same information as a background trader, although it has no private value. The market maker computes $\hat{r}_T$.

The market maker's previous orders are automatically canceled. The market maker then must submit a \emph{ladder} of limit orders, comprising an equal number of BUY and SELL orders centered around the expected final fundamental value $\hat{r}_T$:
\begin{align*}
p_{SELL} \in \hat{r}_T + j \times \xi \\
p_{BUY} \in \hat{r}_T - j \times \xi,
\end{align*}
for \emph{rungs} $j \in \{1, 2, \ldots, K\}$ for $numRungs$ $K$, and $rungSize$ $\xi$.

\subsection{Agent Payoffs}

A market maker in our model has no private value and starts each run with no cash and no inventory. As a result, a market maker's payoff is equal to its cash holdings after its final inventory,$I_T$, is liquidated at price $r_T$. Equivalently, a market maker's total payoff is the sum over its transactions $j$ of the price it receives $p_j$, including the final liquidation as a transaction.

A background trader also starts each run with no cash and no inventory, but it has a private value for each unit it holds in its inventory. A background trader's payoff is the sum of its final cash holdings after liquidation, and the marginal private value of each unit it holds at liquidation time. In other words, a background trader's payoff is the sum over its transactions $j$ of the price it receives $p_j$ and the private value it gains $v_j$, plus the liquidation value of its final inventory, $I_T \times r_T$.

\section{Trading Strategies} \label{strategies}

Many trading strategies have been proposed for the CDA. Past strategies from the market literature have been developed for different purposes and with different assumptions about what parameters of market state agents can sense, so not all are directly comparable to one another.

\subsection{Zero Intelligence}

The Zero Intelligence (ZI) strategy was originally used to show that when agents have private values for shares, the CDA can produce an allocation that nearly maximizes social welfare, even if agent strategies are very simple \cite{gode1993allocative}. By definition, the Zero Intelligence strategy assigns each agent two parameters, $R_{min}$ and $R_{max}$, with $0 \leq R_{min} \leq R_{max}$. Each time a ZI agent arrives at the market, it randomly chooses a surplus to demand by drawing from the uniform distribution over $[R_{min}, R_{max}]$. For example, if the public value of a share is $102$, the agent's private value for a share is $5$, $R_{min}$ is 5 and $R_{max}$ is $10$, the agent might randomly draw a desired surplus of about $7.1$, and submit either a SELL order at price $114.1$ or a BUY order at price $99.9$. The agent's choice of whether to become a buyer or seller is determined by ``coin flip,'' or uniform random draw from the set $\{BUY, SELL\}$.

\subsection{Adaptive CDA Agents}

A succession of adaptive strategies for CDAs has been proposed by various authors. These adaptive policies are meant to perform well across a variety of environments, considering both the parameters of the market such as the volatility of the stock's public value, and the strategies of other agents in the market. In contrast, a particular ZI strategy parameterization may yield an acceptable expected payoff in one market setting, but not in others, and has no way to adapt accordingly.

One early proposal for an adaptive CDA strategy was Zero Intelligence Plus (ZIP), an online learning variant of ZI \cite{cliff1997minimal}. A ZIP agent maintains a memory of its most recent order, including whether it was a BUY or SELL order and its price, as well as whether that order transacted immediately upon submission to the market. The ZIP agent also keeps a factor $\mu$ that is intended to signal if the agent should demand large or small surpluses. Each SELL (BUY) order the agent submits is increased (decreased) from the agent's value per share by a factor of $\mu$. When the agent receives a signal from the market that its last order demanded less surplus than it could have in order to transact, $\mu$ is increased toward 1. In contrast, if an order fails to transact immediately, this may lead the agent to reduce $\mu$ toward 0. Over time, the agent adapts the surplus it demands from each trade to market conditions.

A more sophisticated adaptive CDA strategy is known as GD, after its authors, Steven Gjerstad and John Dickhaut \cite{gjerstad1998price} \cite{tesauro2001high}. The GD agent seeks to outperform ZIP by maintaining, instead of a single $\mu$ value, a history of the most recent orders to BUY or SELL, their prices, and whether they transacted. For each price in the history, the GD agent uses a heuristic to estimate the probability of a new order transacting at that price. When the GD agent arrives at the market, it places an order at the price that maximizes expected profit, computed as the product of the likelihood of transacting and the surplus from a successful transaction.

\subsection{Our Approach}

Our approach assigns each agent a non-adaptive strategy, in the same family as ZI, and unlike ZIP or GD. I use non-adaptive strategies because, for a particular market environment, considering both the parameters of the market and the set of other agents, adaptation may not be particularly useful. (It may be beneficial to select actions based on the time remaining, but this it not really an adaptation.) Our approach, described in Section \ref{egta} on empirical game-theoretic analysis, is to find stable mixtures of non-adaptive strategies, which perform nearly optimally in a given environment. I will not discuss inherently adaptive strategies like ZIP or GD further in this work.

\subsection{Zero Intelligence with Reentry}

A recent study of CDA markets considered the influence of traders called \emph{market makers} on other agents \cite{wah2015welfare}. This study considered parameterizations of the ZI trading strategy. The ZI agent as defined for this study must reenter the market at randomly generated times, at which time it must cancel its previous orders, and can optionally submit new orders.

At each arrival at the market the ZI trader is uniformly randomly assigned an order type for that arrival only, from $\{BUY,SELL\}$. The ZI agent submits an order to buy (sell) at the expected final public value of a share, plus the agent's private value of the share, minus (plus) the desired surplus, drawn uniformly randomly from $[R_{min}, R_{max}]$.

In addition, the ZI agent has a parameter $\eta \in (0, 1]$. If the trader could obtain $\eta$ fraction of the desired surplus, by submitting a \emph{market order} at the current BID or ASK price (best price in the limit order book to sell or buy), the ZI agent will submit a market order instead of a limit order.

For example, suppose a ZI agent has $R_{min} = 5$, $R_{max} = 10$, the expected final share price is 100, and the private value is $5$. Suppose the agent draws a desired surplus of $6$ from $[5, 10]$, the agent draws BUY, and the agent has $\eta = 0.5$. Say the ASK is $101$. The agent's expected surplus from buying at the ASK is $5 - (101 - 100) = 4$, which is greater than $\eta \times 6 = 3$, so the agent submits a market order to buy immediately at the ASK.

\subsection{RL-Derived Strategies}

In this work, I use reinforcement learning to derive strategies for competing against ZI agents in a CDA market. The learning agents are trained either online, using a version of Monte Carlo Tree Search known as Partially Observable Monte Carlo Planning (POMCP) \cite{silver2010monte}, or offline, using an iterative method such as Q-learning or Sarsa$(\lambda)$.

These learning agents, like the ZI agents in our simulation, are given limited information about the state of the market. The learning agents can sense the current bid and ask prices, the current fundamental value of the stock, their own private value, their own inventory, and the number of time steps remaining in the simulation. These agents arrive at the market under the process as the ZI agents.

I have experimented with two settings for the learning agents: FlipKnown, where the agent is randomly assigned to BUY or SELL before at each arrival at the market, and NoFlip, where the agent may choose whether to BUY or SELL at each arrival. Obviously, higher payoffs are expected from the NoFlip agents. The FlipKnown agents are a more fair comparison with ZI agents, however.

\section{Empirical Game-Theoretic Analysis} \label{egta}

Empirical game-theoretic analysis (EGTA) is a method for studying the likely outcomes when agents interact strategically in an environment, but no analytic solution is known for the problem of predicting the agents' behavior \cite{wellman2006methods}. The new work I present in this paper does not employ EGTA directly, but EGTA was used to produce the mixture of other-agent strategies, for our learning agent to compete against. One key aim of this work is to test whether the so-called equilibrium strategy profiles found by EGTA in our market setting are likely to be Nash equilibria, or can be refuted. For this reason I present a description of EGTA here.

Recall that EGTA seeks to produce a reasonable estimate of how a set of self-interested agents might behave in a given environment, based on the assumption that agents are approximately rational, which is to say they approximately maximize their expected utility. While the EGTA methods could be applied to agents with different goals, such as risk-averse agents, what is essential for EGTA is that agents act to maximize some function of their environment in expectation.

EGTA is by definition empirical, and at the heart of the method is a simulator of the environment. The simulator generates sample payoffs for each agent, given an assignment to each agent in the environment of a strategy or policy to play, which I call a strategy \emph{profile}. An agent's strategy, as usual, is a mapping from each possible observation of the environment to an action for the agent to take. An agent's strategy can be either pure, implying that the agent deterministically selects its policy, or mixed, meaning the agent draws its policy from a distribution at simulation start time. Expected payoffs for a policy can be estimated with any accuracy desired, by taking enough samples from the simulator.

EGTA operates iteratively, with an \emph{inner loop} inside an \emph{outer loop}. Within the inner loop, agents are restricted to playing any mixture of a finite set $\mathcal{S}$ of strategies, typically different versions of one parameterized strategy such as ZI. Given the environment's agent count $N$, say $5$ agents, every combination of the available strategies $\mathcal{S}$ is sampled some number of times. Any game that works well in EGTA will be symmetric (payoff distributions depend only on how many agents play each strategy, not which agent does), or at least role-symmetric (each agent is labeled with a role, and payoff distributions are symmetric within each role). So given $N$ agents and $|\mathcal{S}|$ strategies in a symmetric game, there are  $\binom{N + |\mathcal{S}| - 1}{N}$ possible profiles \cite{wiedenbeck2012scaling}. Each of these profiles is sampled some minimum number of times, and the sample mean payoff for each strategy is recorded.

Given an accurate estimate of the mean payoff from every symmetric profile for $N$ agents playing strategies in $\mathcal{S}$, it is possible to compute equilibrium strategy profiles, in the game where agents are restricted to playing mixed strategies over $\mathcal{S}$. Our implementation of EGTA uses a local search method called \emph{replicator dynamics} \cite{taylor1978evolutionary} to search for \emph{role-symmetric mixed-strategy Nash equilibria} (RSNE). This process works by starting from a random point in the standard simplex containing the mixed strategies over pure strategy set $\mathcal{S}$. The current point is then moved according to a hill-climbing rule that increases the chance of playing strategies that are good responses to the current mixed strategy. When replicator dynamics finds an approximate fixed point, that point is an approximate Nash equilibrium. Although replicator dynamics is not guaranteed to discover RSNE if they exist, it tends to work quickly and discover equilibria in our settings in practice.

The \emph{outer loop} of EGTA simply adds new strategies to the set $\mathcal{S}$, before the inner loop begins again. These new strategies might be created by the intuition of the experimenter, or by reinforcement learning against the previous equilibrium.

\subsection{Solution Concepts}

There are many ways of predicting which mixed strategy a set of rational agents would play in a multi-player game, given the expected payoff each agent would receive for every possible strategy profile. These ways are known as \emph{solution concepts}, the most famous of which is the Nash equilibrium. A mixed strategy profile is a Nash equilibrium if no agent can increase its expected payoff by unilaterally switching to some other pure strategy, while other agents hold to the same mixed strategy. A Nash equilibrium is considered a reasonable guess for how rational agents might behave, because it is stable, in that no agent acting alone has an incentive to depart from the equilibrium policy. In a role-symmetric Nash equilibrium, for every role, all agents of that role play the same mixed strategy. I will consider only RSNEs in this work, although there may exist non-role-symmetric Nash equilibria in our selected environments as well.

\subsection{Player Reduction}

Even though we treat the market simulation game as role-symmetric, with all background agents having one role and the market maker another, there are too many profiles to simulate with $25$ background traders and several strategies. For example, with $10$ strategies and $25$ background traders, there are $\binom{34}{25}$ profiles to sample, or \num[group-separator={,}]{52451256}.

%import scipy.misc; scipy.misc.comb(34,25);

We use \emph{player reduction} to reduce a game with many players and strategies to a more manageable size. Ideally, a strategy profile in the reduced game should map to a strategy profile in the full game, where each role produces a similar expected payoff for each agent as in the reduced game.

The reduction method we use is \emph{deviation-preserving reduction} (DPR), approximating a game with $25$ background traders and 1 market maker, by a game with $5$ background traders and $1$ market maker. The \emph{self agent}, if a market maker, controls $1$ market maker in the full game, while each other agent controls $5$ background traders. If the self agent is a background trader, it controls $1$ background trader in the full game, while each other trader agent controls $6$ background traders in the full game. Thus, with 10 background trader strategies, the DPR reduction requires sampling only $|S| \binom{n + |S| - 2}{n - 1}$ profiles, whichin this case is $10 \times \binom{5 + 10 - 2}{5 - 1} = 7150$ profiles \cite{wiedenbeck2012scaling}.

\subsection{EGTA Limitations}

There are multiple reasons one might doubt the correctness or usefulness of equilibria produced by the EGTA process. These stem from the finite strategy set in the inner loop, the non-exhaustive search for equilibria, and the player reduction.

First, consider the restricted strategy set in the inner loop. At best, with error-free estimates of the mean payoff for every inner loop profile and an exhaustive search for Nash equilibria, the inner loop finds all Nash equilibria for a restricted version of the \emph{true game}. In the true game, agents can play any possible strategy, including infinitely many parameterizations of their strategy type, as well as other strategies outside that strategy space. The inner loop of EGTA considers only that finite set of strategies $\mathcal{S}$ that it is given by the outer loop. There could be some superior strategy that has not been sampled.

Next, consider the equilibrium search, where we currently use replicator dynamics. It may be that some equilibria exist that replicator dynamics fails to discover, which produce different outcomes from the equilibria that are found. This would lead to misleading conclusions from EGTA.

Finally, consider the player reduction, which introduces a degree of approximation into the equilibrium search within the inner loop's restricted strategy set. It is possible that each supposed equilibrium found during equilibrium search over payoffs from the reduced game is not an equilibrium in the full game. To see this, suppose that EGTA with DPR finds a mixed-strategy equilibrium where any agent has probability $0.5$ of playing strategy $A$ or strategy $B$, and the player count is reduced from 25 to 5. This means that no agent can benefit in expectation by defecting to strategy $C$, while other agents play this mixed strategy. EGTA has estimated the payoff of alternative strategy $C$ against the equilibrium mixture using only those profiles contained in the DPR reduction, where the other agents play $A$ or $B$ in groups of 6. All profiles where numbers of agents not divisible by 6 play $A$ are missing from the DPR payoff estimates, and if the missing payoffs are much different from the others, then the equilibrium found by EGTA with DPR could be spurious.

In this work, I seek to respond to the first challenge to EGTA, that the strategy set $\mathcal{S}$ in the inner loop is too restricted to give insight into how agents would behave in the true game. I use reinforcement learning over a larger strategy space than $\mathcal{S}$, to determine how much greater an expected payoff agents can earn against an equilibrium mixed strategy from EGTA.

\section{Reinforcement Learning} \label{rl}

\subsection{The RL Problem}

In reinforcement learning, there exists a goal-oriented agent in an environment, where the agent can sense the environment, act in the environment in response to its perceptions, and receive a reward that depends on its actions. The reinforcement learning problem is to direct the agent's actions, conditioned on its observations, in a way that is expected to produce high rewards on average over time. Because the agent learns about its environment over time as a result of choices it has made about which actions to take, the agent faces an explore-exploit trade-off. The agent must balance its need to learn about the payoffs it would receive by taking new actions, with its desire to earn high payoffs by taking the action that currently appears to yield greatest expected return.

Note that in this work I only want to find a nearly-optimal fixed policy for a background trader to use in a particular environment, without regard to payoffs during training. I do not care how high or low the payoff of the learning agent may be during the training process, only how high the payoff is of the best resulting policy. During training, I do need to explore the policy space to evaluate alternative policies which may provide higher payoff than those already sampled. Computational resource limitations require me to perform this search efficiently. But there is no penalty in our setting for low mean payoffs during the agent's training time.

\subsubsection{POMDPs}

I consider here the case where the agent's environment evolves as a partially observable Markov decision process (POMDP). The agent's environment has a set of states $S$, a set of observations $O$, and a set of actions $A$ the agent can take, all of which are known to the agent before training. The environment has a transition probability distribution $T(s,a,s')$ which maps any state-action pair $(s,a) \in S \times A$ to a probability distribution over next states $s' \in S$, corresponding to the probability that the system will transition to state $s'$ if the agent plays $a$ in state $s$. The existence of this transition function $T$ implies that a POMDP evolves according to an underlying MDP, which the agent cannot observe directly. There is also an observation distribution $O(s,a',o)$ which maps any state-action pair $(s,a')$ to a probability distribution over observations $o \in O$ that the agent may receive when the agent takes action $a$ and transitions to state $s'$. There is a reward distribution $R(s,a)$ which maps each state-action pair to a continuous probability distribution over the reals, representing the immediate (marginal) payoff distribution when the agent plays action $a$ in state $s$. Finally, the model has a probability distribution $W(s)$ over initial states of the environment. I assume that there is no discounting of rewards, so $\gamma = 1$. Total payoffs in our market simulator do not tend toward infinity, because simulations are discrete-time and bounded in duration.

\subsubsection{POMDP-Solving Approaches}

One approach to learning in a POMDP setting is to treat the POMDP as an MDP in belief space (or, equivalently, in history space), and to use an MDP solution method on the belief MDP. In this approach, the learner keeps track of the complete history $H$ of actions and observations in each training playout. A history has the form $H = (o_0, a_0, o_1, a_1, o_2, \ldots)$, containing an initial observation, the agent's response, the next observation, and so on. The learner treats a sequence of alternating actions and observations as a ``state'' in the belief MDP. I will use this approach in our POMCP learner, described in more detail below. The benefit of the belief MDP method is that it offers guaranteed convergence to the optimal solution with sufficient sampling, using conventional methods like Q-learning. Unfortunately, using histories as states creates a combinatorial explosion in the state space, leading to poor generalization between similar histories, slow learning, and inefficient memory use.

In a POMDP setting, another approach for an RL learner is to learn a restricted-memory policy, in the extreme case a memoryless policy. When learning a memoryless policy, the learner essentially treats the POMDP as if it were a fully-observable MDP, and handles the observations in $O$ of the POMDP as if they were actually states of an MDP. The distinction between observations and states is that in a POMDP, a certain observation $o \in O$ can be sensed in different underlying states $s \neq s'$, with different transition probabilities and payoffs.

The downside of treating a POMDP as an MDP and using conventional techniques like Q-learning or Sarsa is that it is not clear that these methods will converge to the best deterministic memoryless policy, no matter how many samples are taken \cite{loch1998using}. Moreover, in some degenerate problems, the best deterministic memoryless policy (such as what Q-learning or Sarsa searches for) may be arbitrarily worse than the best stochastic memoryless policy \cite{singh1994learning}. In fact, even if the transition and payoff functions are known, it is NP-hard in general to determine the optimal memoryless policy \cite{littman1994memoryless}. The benefit of approaching the problem in this way, however, is that many practical POMDPs can be solved effectively with classical RL techniques treating observations as states; this approach is simpler to implement than POMDP-specific methods like Monte Carlo Tree Search; and memoryless policies are easier to interpret than policies conditioned on histories.

A memoryless policy $\pi$ maps the observation space $O$ to the action space $A$. In a POMDP with absorbing states, stochastic start states, and no discounting, the learning agent searches for the policy $\pi$ that maximizes expected return:
\begin{align*}
\pi \in O \to A : \textrm{Maximize } \mathbb{E}_{s_0 \sim W} \bigg( \sum R \Big(s_i, \pi(o_i) \Big) \bigg),
\end{align*}
where $s_i$ for positive $i$ is distributed based on the transition function $T$, and observation $o_i$ is distributed based on observation function $O$. A belief-state policy is similar, but it maps histories $H$ to actions, where a history is any alternating series of observations in $O$ and actions in $A$ that begins and ends with an observation.

\subsection{Our RL Environment}

Our learning agent is a background trader in a simulated stock market with one stock, as described above. The agent attempts to learn a policy mapping its observations of the state of the market to trading actions, in the form of orders to submit.

Each environment I consider is derived from a game discussed in a recent study of market makers in a CDA \cite{wah2015welfare}. To produce a POMDP from a game, I simply select one agent in the background trader role to become the self agent, and fix the strategies of all other agents. The result is a POMDP where the self agent has the same action set as the corresponding agent in the original game, and the world evolves between the agent's decisions as the original game would, with the other agents playing their assigned strategies.

\subsubsection{State Set}

A state $s \in S$ of our POMDP encodes the complete state of the simulated market, excluding random seeds used by the simulator implementation. This state has an integer number of time steps remaining in $\{0, 1, \ldots, T\}$. It has a set of orders in the limit order book, each labeled BUY or SELL, owned by a particular agent who placed the order, with a positive integer price. It has the pure strategy being used by each other agent, as drawn from the other-agents mixed strategy for the appropriate role. It has the private value vector of each agent, the integer-valued inventory $I$ of each agent (which must sum to 0 over the agents), and the integer-valued cash of each agent (which also sums to 0). Finally, the state has the current non-negative real fundamental value $r_t$ of the stock.

\subsubsection{Action Set}

An action $a \in A$ of the self agent specifies what order to place when the agent arrives at the market, after clearing its previous orders. The order can be to BUY or SELL (unless running in a mode where the agent is assigned an order type for each arrival). In addition, the agent can choose to play NOOP, not placing any order.

Most importantly, the agent can choose how much surplus to demand. The surplus demanded is defined to be, for a BUY order, the total value of a share to the agent minus the order price. For example, if the agent places a BUY order at $100$, the expected final value of a share $r_T = 98$ and the private value of the next share obtained is $5$, the total value of a share acquired by the self agent is $103$, so the agent is demanding a surplus of $3$. The surplus demanded for a SELL is defined similarly, with the difference reversed.

The agent is allowed to choose a price for its order by selecting from a finite set of surpluses to demand, for example, $\{30, 60, 120, 240, 360\}$. Given a  surplus to demand, the agent will compute the order price from the expected final fundamental value and change in private value from gaining or losing one share, then place the appropriate order.

\subsubsection{Observation Function}

An observation $o$ in our market POMDP is merely a filtered view of the current state $s$ of the market. The self agent is not able to observe aspects of the market such as the pure strategy of the other agents, their private values, their inventories, or which agents submitted which orders in the past. The self agent receives an observation only when it arrives at the market, meaning when it is allowed to take an action, according to its exponential inter-arrival time process.

Given a market state $s$ when the self agent arrives at time $t$, the agent receives an observation $O(s)$ that includes the number of time steps left $T - t$, the current fundamental value $r_t$, the current best BID and ASK price in the limit order book (if any), the self agent's inventory $I$, the self agent's cash, and the self agent's private value vector.

\subsubsection{Transition Function}

When the self agent takes an action $a$ in market state $s$, the transition function generates the next state $s'$ of the market when either the agent arrives again, or the simulation reaches time $T$ and terminates. The next arrival time of the self agent, as for all agents, is sampled from an exponential inter-arrival time distribution, which is rounded to an integer tick value, and added to the previous arrival time, as described earlier. If this next arrival time is greater than $T$, the simulation enters a terminal state $s'$.

The parameters of the next state $s'$, besides the time step $t'$ addressed above, are determined by our market simulator. While the market simulator runs between POMDP states, it generates arrivals for the other agents, which cancel and submit orders according to their predetermined strategies. Some of these orders might match and clear, leading to transactions, which change the associated agents' inventory and cash variables. Of course, the fundamental value evolves with each time step according to its random process.

\subsubsection{Reward Distribution}

The self agent is a background trader, so its total return during a simulation run must equal the payoff for a background trader described earlier: The return is the background trader's cash after liquidation, plus the net private value of shares held at liquidation time. It is helpful for RL methods to realize rewards before the end of a simulation, however, so in our POMDP formulation I give the self agent intermediate rewards that are guaranteed to sum to the desired total return.

I define the reward to the self agent to be $0$ in its initial state, when it has zero inventory and zero cash. Otherwise, suppose the self agent takes action in state $s_t$ and arrives at next state $s_{t+k}$, $k > 0$, which could be the terminal state $s_T$. The agent's cash changes from $c_t$ to $c_{t+k}$, its inventory changes from $I_t$ to $I_{t+k}$, and the expected final fundamental value of a share changes from $\hat{r}_t(T)$ to  $\hat{r}_{t+k}(T)$. Then the intermediate reward is:
\begin{align*}
R_{t+k} = \Big(I_{t+k} \times \hat{r}_{t+k}(T) - I_t \times \hat{r}_t(T) \Big) + \Big( c_{t+k} - c_t \Big).
\end{align*}
In the terminal state $s_T$, the self agent receives an additional intermediate reward equal to the cumulative private value of its inventory $I_t$, which I denote $v_T$. Observe that summing the rewards over all the self agent's arrivals yields, as desired:
\begin{align*}
\sum R = (I_T \times r_T + c_T) + v_T,
\end{align*}
because the initial inventory $I_i$ and cash $c_i$ are zero. Therefore, the sum of the intermediate rewards in our formulation equals the background trader's total surplus.

\subsubsection{Initial State Distribution}

The initial state distribution $W(s)$ is generated similarly to the next state, through the operation of the market simulator. The self agent's initial arrival time is distributed according to the same exponential inter-arrival time process $\textrm{Exp}(\lambda_{BG})$ as its subsequent arrival times. In some cases, the self agent does not arrive at all, and its initial observation is of a terminal state, with reward $0$. Otherwise, the initial state corresponds to the self agent's first arrival at some time $t \leq T$. The state of the market is determined, as usual, by the market simulator, which has generated the arrivals and actions of the other agents, as well as the fundamental value process up to $r_t$.

\subsection{Market Simulator Refactoring for RL}

The stock market simulator used for experiments presented here, as well in prior work on latency arbitrage \cite{wah2013latency} and market making \cite{wah2015welfare}, was not originally designed for use with reinforcement learning. I call this simulator MarketSim, for market simulator. The online reinforcement learning methods I apply in this work, such as Q-learning, Sarsa$(\lambda)$, and POMCP, require a generative model of the environment, from which the learner can select which action $a \in A$ to sample from the current state $s$ before the simulation proceeds.

Before this project was begun, MarketSim had no facility for halting simulation progress when a self agent arrives at the market. I needed to extend the simulator so that in a new operation mode, a self agent could be designated, such that each time the self agent arrives at the market, the simulator pauses to wait for external code to specify the self agent's next action. In addition, the simulator would need to specify a payoff for the self agent after this action was taken, such that the sum of the payoffs would equal the self agent's total profit during the simulation run.

\subsubsection{A New RL Library}

For this paper, I have implemented a general-purpose library of reinforcement learning methods in Java, called EqRegret (because it can be used to evaluate the game-theoretic regret of a proposed equilibrium). The library allows a programmer to create domain-specific classes representing \texttt{State}, \texttt{Action} and \texttt{Observation} objects in their POMDP or MDP.

The centerpiece of the library is the \texttt{SamplingOracle} class, which the user must extend to provide the generative model of an environment, which various learning methods can be applied to. A \texttt{SamplingOracle} extension provides a method \texttt{generateSample}, which takes a \texttt{State} $s$ and an \texttt{Action} $a$ as arguments, and returns a \texttt{Sample} object that contains the next \texttt{State} $s'$, the reward, and the next \texttt{Observation} $o$. The method \texttt{sampleInitialState} returns an initial \texttt{Sample} object, with associated \texttt{State}, \texttt{Observation}, and reward.

The library implements several popular RL methods, which can be applied generically to any problem specified as library objects. Among the methods that have been implemented and tested for this work are Q-learning (with and without tile coding or experience replay), Sarsa$(\lambda)$ (with and without tile coding), and POMCP. Each of these methods has been extensively validated on example problems from the literature, as described in section \ref{experiments}.

\subsubsection{MarketSim and EqRegret}

In our MarketSim code for RL, the \texttt{State} object is a wrapper for a \texttt{Simulation} object, which holds the complete state of the market simulator for some state $s$. When the \texttt{SamplingOracle} needs to generate the next state $s'$ after the self agent plays an action $a$, the inner \texttt{Simulation} is used to sample $s'$. One difficulty with this approach is that the original \texttt{Simulation} $s$ must not be mutated during sampling of the next sample $s'$. If the original state $s$ were altered, it would not be possible to take additional samples from $s$, as needed by Monte Carlo Tree Search. Our solution to this problem is to use Java's native serialization code and \texttt{ObjectOutputStream}/\texttt{ObjectInputStream} to perform a deep copy of the complete market state in the \texttt{Simulation}, then generate the next state $s'$ by mutating the deep copy.

A second challenge with the wrapper approach is that the \texttt{Simulation} object stores its own random seeds, and those seeds are retained by a deep copy. As a result, all runs from a deep copy of a particular \texttt{Simulation} state $s$ would produce the same results, instead of sampling randomly from a distribution as desired. To fix this problem, I reset the random seed of every pseudorandom number generator in the model, in each new copy of a \texttt{Simulation}. A similar issue is that MarketSim generates certain events before the time step in which they are removed from the priority queue, such as the arrivals of other agents, or the evolution of the fundamental value. I wrote code to clear these ``future events'' when the self agent arrives, and to generate them again when the simulation is restarted after a pause, from the appropriate conditional distribution. If I had not done this, future fundamental values and other agents' next arrival times would have been identical, each time the simulation was run forward after a pause.

Before this work, the MarketSim simulator had no facility for halting execution when a certain agent arrives at the market, then restarting progress after the agent has selected its action. MarketSim's discrete-time simulation operates with a priority queue of events, each of which has a timestamp that determines its priority. I modified the \texttt{Scheduler} class to inspect each event as it is dequeued, to check if it is an arrival of the self agent. When the self agent arrives, the simulation time is set equal to the arrival time, and a flag is set that the simulation should halt. External code is presented with the current state of MarketSim, and it returns the next \texttt{Action} of the self agent. An event corresponding to this action is executed immediately, and then the \texttt{Scheduler} is set in motion to generate the next state.

\subsubsection{Validating the Market Simulator Changes}

I have performed extensive unit tests to verify that MarketSim performs in the same way through stops and starts in RL-mode, as it does in the original mode that was used in prior work. For example, I performed statistical tests over many simulation runs to verify that the self agent and the other agents arrive at the market the correct number of times on average, based on their exponential interarrival parameters. I have tested that the fundamental value evolves over time correctly, whether MarketSim is run in RL-mode or not. Most importantly, I have verified that the expected payoff of running a particular strategy is the same for the self agent and the other agents, in RL-mode or in the original mode.

\subsection{RL Challenges}

The MarketSim environment poses several challenges for a reinforcement learner, including partial observability, a large, continuous state space, noisy payoffs, and time-consuming sample generation.

Partial observability arises in our RL problem from the agent's limited observations of the state of the underlying market POMDP. As described earlier, the agent is not able to observe the private values, inventories, or assigned strategies of other traders. The agent also cannot see orders in the limit order book beyond the best bid and ask. As a result, there is perceptual aliasing, in that multiple underlying states can have the same appearance to the self agent. For example, if the agent observes that no orders currently exist in the limit order book, the agent cannot tell whether the other agents are biased toward buying or selling, based on their private values.

The state space of our MarketSim POMDP is extremely large. In a typical simulation run for this paper, the simulation runs for $1000$ time steps, with fundamental values ranging from \num[group-separator={,}]{80000} to \num[group-separator={,}]{120000}, and $25$ background traders each with its own private value vector of $20$ entries, as well as its own inventory and cash variables. A reinforcement learner must use generalization over the infinitely many possible states in order to learn. I approach this problem by discretizing states into buckets, using a vector of threshold values along each dimension such as fundamental value, and letting learners estimate the payoff for each group of similar states.

\begin{figure*}
\centering
\includegraphics[width=0.7\columnwidth]{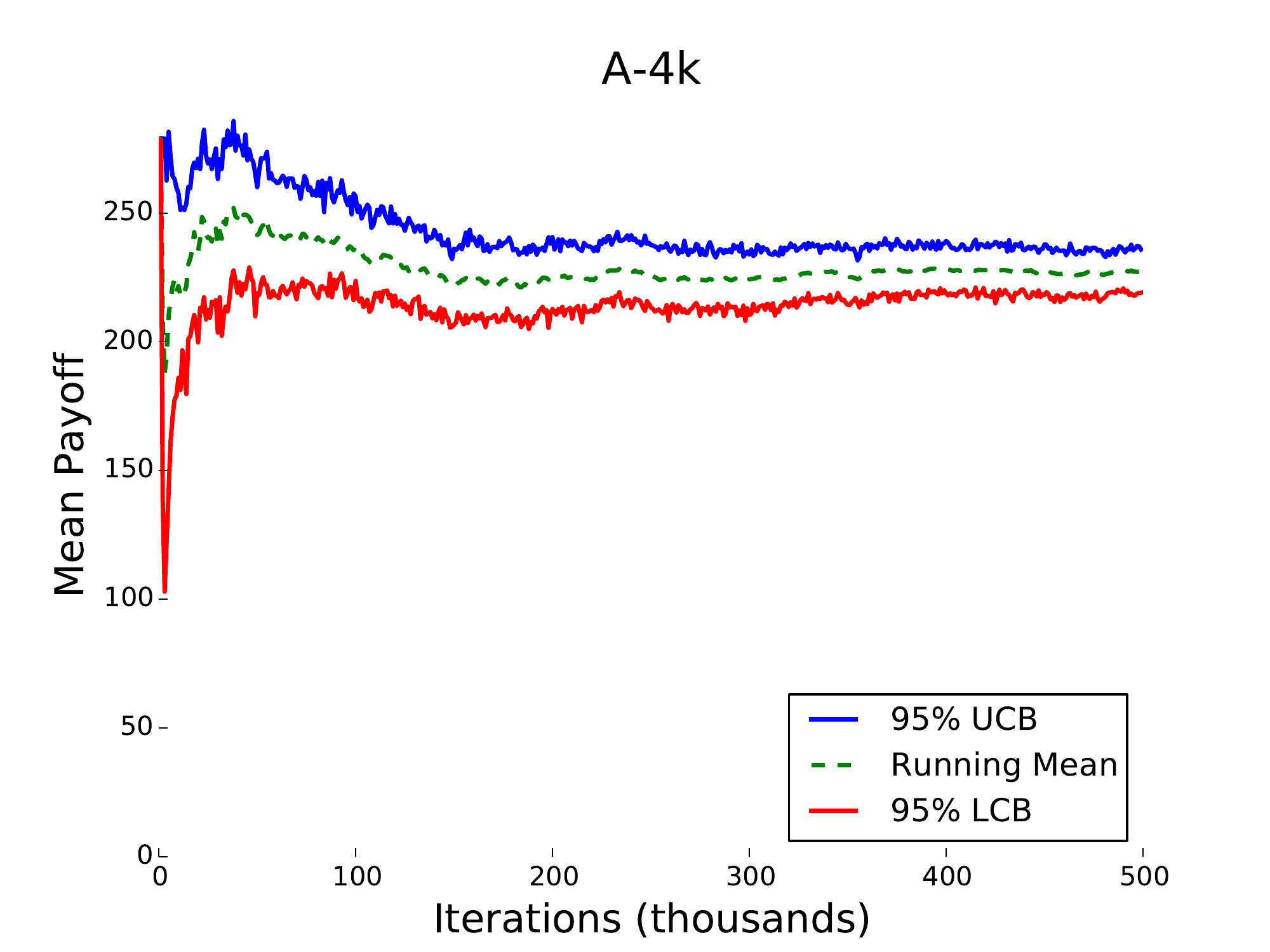}
\caption{Many payoff samples are needed to obtain a precise estimate of the true mean payoff of a policy. Sample payoffs are from MarketSim, for EGTA's equilibrium strategy in the A-4k environment. The dashed line shows the running mean of self-agent payoffs observed so far. The solid lines show a bootstrap $95\%$ confidence interval for the true mean, based on preceding observations only.}
\label{fig:runningMean}
\end{figure*}

An RL agent in the MarketSim environment seeks to choose a policy that maximizes its expected payoff, but the payoffs for any policy are quite noisy. Sources of payoff variation for a CDA trading policy include all the sources of randomness in the market model, such as agent arrival times, the fundamental value process, and the agents' private value vectors drawn at start time. As an example of how noisy the payoffs are, consider the environment $A-1k$ (described in Section \ref{experiments}), where an example learned policy has sample mean payoff $61.48$ after \num[group-separator={,}]{100000} simulations. The range of payoffs observed is from about \num[group-separator={,}]{-17000} to \num[group-separator={,}]{24000}. The sample standard deviation of the payoffs is $1355.69$, and a bootstrap $95\%$ confidence interval for the true mean, based on $1000$ resample means form the \num[group-separator={,}]{100000} simulated payoffs, ranges from about $53.5$ to $69.7$. Note that the range from $53$ to $69$ contains the expected payoff of every constant-action policy in the A-1k RL action space. As shown in Figure \ref{fig:runningMean}, the sample mean payoff for a MarketSim policy does not become reliable until after a few hundred thousand samples. This indicates that a large number of payoff samples (hundreds of thousands or millions) is needed to estimate the value of a MarketSim policy with sufficient precision to make informed decisions.

The last key challenge for RL in the MarketSim environment is costly sample generation. In many POMDP environments where RL has proven highly successful, it is inexpensive in time and memory to generate many sample observations of the next state, observation, reward, given a prior state and an action. POMDPs that are cheap to sample include board games like Battleship and Amazons, as well as grid worlds like the Sailing stochastic shortest path problem. The MCTS approach has been highly successful on all of these problems, as instantiated in Upper Confidence Bounds on Trees (UCT) or POMCP \cite{silver2010monte} \cite{kocsis2006improved}. MDPs where the next state is cheap to sample, such as Sutton's gridworld, have long been easy to solve using Q-learning, Sarsa$(\lambda)$, or other classical approaches \cite{loch1998using} \cite{littman1994memoryless}.

The MarketSim environment's high payoff variance means that many samples must be taken to estimate expected payoffs precisely, and the complexity of the market simulator means that each state sampling operation is slow, compared to state updates in a board game like Battleship. For example, running on a single core of a server with Intel Xeon 3.2 GHz processors, a set of \num[group-separator={,}]{100000} simulation runs of the MarketSim environment requires about $17.5$ minutes in the A-1k environment (with $1000$ time steps per run), $44.4$ minutes in the B-1k environment (due to a higher arrival rate for background traders), and $72.0$ minutes in the A-4k environment (with $4000$ time steps per run). For comparison, \num[group-separator={,}]{100000} runs of a typical learned policy in the Sailing[20,20] environment \cite{kocsis2006bandit}\cite{peret2004line} requires only $1.9$ minutes. \num[group-separator={,}]{100000} runs of learned policies in the Sutton's gridworld environment \cite{littman1994memoryless} \cite{sutton1990integrated} requires only $3.6$ seconds.

\subsection{RL Methods}

\subsubsection{Q-Learning and Sarsa}

I have applied two classical methods for learning in an MDP setting, Q-learning and Sarsa. Both of these methods are iterative, online algorithms, meaning that they update their predictions continually as new samples are produced by the simulator.

Both Q-learning and Sarsa attempt to learn the value of taking any action $a \in A$ in any state $s \in S$. In an MDP setting, there always exists an optimal policy that is deterministic and memoryless, which corresponds to taking the action that maximizes the expected total of the immediate reward plus the discounted value of the succeeding state \cite{littman1994memoryless}.

I say that the $Q$-value of a state-action pair $(s,a)$ under the optimal policy $\pi^*$ is the expected return for playing action $a$ in state $s$, given that one will follow the optimal policy thereafter \cite{watkins1992q}:
\begin{align*}
Q^*(s,a) = \mathbb{E}\bigg( R(s,a) + \gamma \sum_{s'} T(s,a,s') \max_{a'} Q^*(s',a') \bigg).
\end{align*}
From this Bellman equation, I can derive the update rule for Q-learning. Suppose that I have, for every state-action pair $(s,a)$, an estimate of the $Q$-value and a count $k$ of observations I have seen. Given a transition from previous state $s$, taking action $a$, to new state $s'$ and receiving reward $R$, I can update the estimated $Q$-value of $(s,a)$ as follows:
\begin{align*}
Q_{i+1}(s,a) \leftarrow (1 - \alpha_i) \times Q_i(s,a) + \alpha_i \times \Big(R +  \gamma \max_{a'} Q_i(s',a') \Big),
\end{align*}
where $\alpha \in (0, 1)$ is the learning rate, which decreases toward zero with increasing visit count $k$.

In their seminal paper on Q-learning, Watkins and Dayan showed that in an MDP setting with bounded rewards, each $Q$-value converges to its correct value under the optimal policy $\pi^*$ with probability $1$ \cite{watkins1992q}. The conditions for this are that each state-action pair be visited infinitely many times in the limit, which may be ensured by using an $\epsilon$-greedy exploration strategy (and having only reachable states). Also, that the sum of the weights $\alpha_i$ for each state-action pair diverge to infinity, while the sum of squared weights converges for each state-action pair. This can be ensured by using $\alpha = \frac{1}{k}$ as the learning rate, or simply tracking the sample mean payoff for each state-action pair.

I note that Q-learning need not converge in probability to the $Q$-value of the best memoryless policy in a POMDP, with the observations treated as states. Similarly, when Q-learning is applied to an MDP with its state space discretized into buckets, there are no guarantees of convergence to the best memoryless policy. Therefore, I cannot guarantee convergence with Q-learning for the discretized version of our MarketSim setting. Moreover, it would be unfeasible to apply Q-learning to the belief MDP of our discretized MarketSim, due to the very large number of histories that are visited with non-negligible frequency.

Sarsa is an online RL method like Q-learning, which is called \emph{on-policy}, because it estimates the $Q$-value of a state-action pair on the assumption that the current Sarsa policy will be taken thereafter (including its exploration mode), rather than the greedy policy (as in Q-learning). The Sarsa update rule takes a transition including the old state $s$, the action taken $a$, the reward $R$, the next state $s'$, and also the next action $a'$ that will be taken by the current explore-exploit Sarsa policy \cite{sutton1998reinforcement}:
\begin{align*}
Q_{i+1}(s,a) \leftarrow (1 - \alpha_i) \times Q_i(s,a) + \alpha_i \times \Big(R + \gamma Q_i(s',a') \Big).
\end{align*}

\subsubsection{Eligibility Traces}

Q-learning and Sarsa are not as sample-efficient as they could be, in that each sampled transition $(s,a,R,s',a')$ is used only once for leaning, to update $Q(s,a)$. In environments where it is expensive to acquire a transition, such as MarketSim, it is worthwhile to apply the information in each sample reward $R$ and next state-action pair $(s',a')$ to more than just the immediately preceding state-action $(s,a)$. I can use \emph{eligibility traces} to propagate credit for the intermediate reward $R$ backward over previous actions in the playout.

Each simulation run during RL training generates a sequence of the form $(s_0, a_0, R_0, s_1,a_1,R_1, s_2,$ $a_2,\ldots)$. When the transition $(s_1, a_1, R_1, s_2, a_2)$ occurs, I can assign some credit for reward $R_1$ to the previous transition, where the agent took state-action $(s_0, a_0)$, appropriately down-weighted by the discount factor $\gamma$.

Sarsa$(\lambda)$ assigns an eligibility trace weight $\eta$ to each state-action pair, which is equal to $0$ at the beginning of a playout for all pairs. When a state-action pair $(s,a)$ is visited, $\eta(s,a)$ is assigned to be $1$. At each subsequent transition where $(s,a)$ is not visited, $\eta(s,a)$ is decreased by a factor of $\gamma \lambda$, for $\lambda \in (0, 1)$. In our experiments, I use $\lambda = 0.9$ and $\gamma = 1$. The value update for Sarsa$(\lambda)$ on each transition $(s,a,R,s',a')$ is applied to all state-action pairs $(s'',a'')$ with non-zero $\eta$ \cite{loch1998using}:
\begin{align*}
Q_{i+1}(s'',a'') \leftarrow \Big(1 - \alpha_i \eta(s'',a'') \Big) \times Q_i(s'',a'') + \alpha_i \eta(s'',a'') \times \Big(R + \gamma Q_i(s',a') \Big).
\end{align*}

To see why eligibility traces can accelerate learning, consider a maze environment where the agent seeks the shortest path to a goal from the start state $s_0$. Partway through training the agent might have found only long paths where it first moves upward, causing the $Q$-value of $(s_0, UP)$ to be poor. In Q-learning or Sarsa without eligibility traces, if the agent later finds a short path to the goal that begins by moving upward, this will immediately improve the $Q$-value estimate for the state just before the goal along the successful path, but it will not propagate the new information back to the initial state at first. Each time the agent follows the successful path again, the favorable $Q$-value will be back-propagated only one transition farther along the path toward the start state. With eligibility traces, this back-propagation takes place immediately, without the need for collecting more samples.

In addition to making learning faster in some cases, some authors have proposed that eligibility traces may increase the stability of classical RL methods like Q-learning and Sarsa in settings like POMDPs where they may otherwise fail to converge to a locally optimal policy in their policy space \cite{loch1998using}.

\subsubsection{Experience Replay}

Besides eligibility traces, another method for improving the sample efficiency of RL is \emph{experience replay}, in which every sampled transition $(s,a,R,s',a')$ is stored in memory, and selected samples are replayed as if they were being experienced again. It may appear odd to perform a $Q$-value update based on the same sample more than once, but experience replay affords similar benefits to eligibility traces for accelerating RL. Returning to the example from the previous section on searching for the shortest path to a goal in a maze, I can see that after a shorter path is discovered which begins $(s_0, UP)$, subsequent replays of preceding steps along this shorter path can propagate the newly discovered payoff back toward the initial state \cite{adam2012experience}.

A typical implementation of experience replay stores past transitions $(s,a,R,s',a')$ in a drop-out queue with its maximum size set as a parameter, perhaps \num[group-separator={,}]{10000} samples. One reason to use a drop-out queue, which automatically discards the oldest samples when the queue is too full to accept a new one, is that memory is limited. Another reason for the drop-out queue is that old samples are distributed according to an earlier policy of the agent with outdated beliefs about the best action to take. Therefore, it may be better not to resample very old transitions \cite{lin1993reinforcement}. Drop-out queues for experience replay have been successfully employed in recent work on Deep Q-Learning, which uses Q-learning combined with a neural net to approximate the $Q$-function, in the Atari games domain \cite{mnih2013playing}.

To draw samples from the memory of past transitions, it is common practice to draw some fixed number $numSamples$ uniformly randomly from memory, after every $runsBetween$ samples from the environment. These values are set as parameters \cite{adam2012experience}. The originator of experience replay, Long-Ji Lin, recommended a variation that replays experiences sequentially in reverse order, although I do not adopt this method here \cite{lin1992self}. In our experiments, I kept a memory of \num[group-separator={,}]{40000} transitions, resampling $numSamples = 4000$ of these every $runsBetween = 1000$ runs, so that a typical sample might be replayed $4$ times. 

\subsubsection{Tile Coding}

The observation space $O$ of MarketSim is large and continuous, so I have discretized it into tiles before applying our RL methods. Each tile occupies a hyperrectangle in the underlying observation space, being bounded above and below along each dimension by threshold values. Every point $o$ in the observation space $O$ falls within exactly one tile, or in other words, the set of tiles form a partition of $O$.

A \emph{tiling} is any set of tiles that partitions its space, such as the observation space $O$ or state space $S$. When just a single tiling is used, a reinforcement learner has limited ability to generalize from an observation at a point $o$ to nearby points $o'$ which are located in a different tile but have similar values. \emph{Tile coding} remedies this problem by using multiple tilings at once, so that nearby points will be in the same tile in at least some tilings. Each tiling is a different partition of the state space, and learning is performed simultaneously on all the tilings, with results from each tiling being propagated to the others \cite{sherstov2005function}.

In tile coding with multiple tilings over a state space $S$, any point $s$ in $S$ falls within exactly one tile in each tiling. Here I will explain the modified $Q$-value update for Q-learning with multiple tilings, but the update for Sarsa is similar. Each tiling maintains its own set of $Q$-value estimates. The $Q$-value update for a state-action pair $(s,a)$ is applied to the corresponding tile $(t_0, t_1, \ldots, t_m)$ from each of the $m$ tilings. For a transition $(s,a,R,s')$, the value $Q(t_j,a)$ is updated as:
\begin{align*}
Q_{i+1}(t_j,a) \leftarrow (1 - \alpha_i) \times Q_i(t_j,a) + \alpha_i \times (R + \gamma \max_{a'} Q_i(s',a')).
\end{align*}
I compute $Q(s,a)$ as the mean over corresponding tiles $t_j$ of the $Q$-value estimate for that tile:
\begin{align*}
Q(s,a) = \frac{1}{m} \sum_{j=1}^m Q(t_j,a).
\end{align*}

In our experiments, I have used either 1 tiling, or 3 tilings.

\subsubsection{Monte Carlo Tree Search}

So far I have described methods such as Q-learning and Sarsa, which converge to the optimal policy under mild conditions in an MDP, but which may not do so when applied to a POMDP, treating its observations as if they were states. I have already ruled out applying Q-learning or Sarsa to the belief MDP of the MarketSim POMDP, due to the large number of histories that are encountered with similar frequency. When too many histories are reached quite often during playouts of an MDP,it becomes unfeasible to sample enough payoffs for each such history to estimate $Q$-values accurately.

An alternative approach is Monte Carlo Tree Search (MCTS). MCTS is an online algorithm that, when given the current state $s$ of an MDP, returns an anytime estimate of the best legal action $a$ to take from that state.

MCTS uses a generative model of the MDP to build a tree that represents possible future states of the MDP, with the current state $s$ at the root. Each node in the tree holds samples of the MDP after a particular history $H$ of states and actions, for example $(s,a_1,s_2,a_2,s_3)$. A node stores how many times each allowable action $a$ has been sampled from the node, along with the sample mean return from taking the action $a$. The MCTS method will have been given a budget of a certain number of nodes to produce in the tree, or any amount of time to run before choosing the action to return.

MCTS proceeds by iteratively running the generative MDP model forward from the current state $s$ until reaching a node in the tree that has an action that has not been sampled yet, and sampling that action, to produce a new node in the tree. MCTS uses a \emph{tree policy} to decide which action of which expandable node it will sample next. After creating the new node for the iteration, MCTS follows a \emph{rollout policy} to move the MDP along to a terminal state or beyond the discount threshold. After a terminal state is reached, the reward is backpropagated to the nodes in the chosen path through the tree \cite{browne2012survey}.

When MCTS has run until the limit of its budget, it returns the greedy action from initial state $s$ that has yielded the highest mean return. Other variants of MCTS besides the one described here have been used, such as returning the most-visited action, but in this work I have simply used the action with highest mean payoff \cite{browne2012survey}.

A key difference among MCTS methods is the design of the tree policy. Before I introduce POMCP, an MCTS approach that has succeeded in POMDPs, I will discuss a theoretically sound tree policy for MCTS on MDPs called Upper Confidence Bounds on Trees (UCT). UCT is the tree policy used by POMCP, so it is relevant to discuss it briefly here.

UCT is a form of MCTS that recursively applies the Upper Confidence Bound (UCB) bandit arm selection method as its tree policy. In the multi-armed bandit problem, a learning agent must repeatedly select one of $K$ arms of a slot machine to pull, each of which has its own iid, bounded-value probability distribution of rewards. The agent seeks to maximize its expected return over any number of pulls.

The UCB1 policy says that the agent should pull the arm $i$ that maximizes:
\begin{align*}
\textrm{UCB}(i) = \bar{R}_{i,n_i} + \sqrt{\frac{2 \ln n}{n_i}},
\end{align*}
after $n$ total samples and $n_i$ samples of any arm $i$. It can be shown that this policy yields an upper bound on regret (relative to pulling the best arm every time) that grows as the logarithm of the number of pulls $n$, which is optimal up to a constant factor \cite{auer2002finite}. UCB1 produces an optimal balance between exploitation, represented by the term $\bar{R}_{i,n_I}$ which encourages sampling the arm of highest sample mean, and exploration, represented by the term $\sqrt{\frac{2 \ln n}{n_i}}$ which encourages sampling the least-sampled arm \cite{browne2012survey}.

UCT is an MCTS variant that uses UCB1 recursively as its tree policy. Starting from the current state $s$, UCT selects the action of highest reward upper-confidence bound. If the action has not been sampled yet, a new node is added to the tree for that action and the rollout policy applied thereafter. Otherwise, UCB1 is applied recursively for the node that succeeds the chosen action. It can be shown that the probability of UCT selecting a sub-optimal action approaches zero as the number of playouts approaches infinity \cite{kocsis2006bandit} \cite{browne2012survey}. Moreover, UCT performs well on some challenging fully-observable domains, such as the Amazons game \cite{kocsis2006improved}.

\subsubsection{POMCP}

MCTS methods for MDPs, such as UCT, cannot be applied directly to a POMDP while maintaining convergence guarantees, for the same reason that Q-learning and Sarsa lose their convergence guarantees when applied to a POMDP treated as an MDP with observations as states. The same observation may be seen for multiple underlying states, creating the problem of perceptual aliasing. Silver and Veness proposed Partially Observable Monte Carlo Planning (POMCP) as an effective way to apply UCT to POMDPs \cite{silver2010monte}.

POMCP builds a tree representing the belief MDP of its POMDP, by mapping each history of actions and observations $H = (o_0, a_0, o_1, a_1, \ldots, o_k)$ to a node. This is feasible for POMCP to do, even though it was not for Q-learning or Sarsa, because POMCP estimates the values only for an asymmetric subtree of the game tree, with the current history at the root. As with UCT, each node keeps track of the number of times each subsequent action has been tried, along with its sample mean return. Also like UCT, POMCP uses the UCB1 tree policy recursively to decide which node to expand, by trying an action that has not been sampled yet and forming a new child node.

The key innovation of POMCP is its method for sampling future rewards and next states, from given history $H$. Note that the generative model of the POMDP returns a next state, observation, and reward for a state-action pair, not an observation-action pair. It is necessary to sample states somehow from the probability distribution over current states $s$, given the POMDP's initial state distribution $W(s)$, conditioned on the history $H$ that has been observed so far. Silver and Veness have shown that the POMCP converges in probability to the value of each node in its tree under the optimal policy, as the number of samples approaches infinity \cite{silver2010monte}.

POMCP samples from the current state distribution by storing in each node a copy of every underlying state $s$ of the POMDP by which that node was reached. If fewer than some parameter $minStates$ matching states have been sampled, POMCP uses rejection sampling to derive more matching states. That is, beginning from the initial state $o_0$, POMCP will draw samples repeatedly until an observation $o$ matching that observed is found; then it takes the next action $a_0$ from the history, until the following observation matches the target history; and so on, until enough samples have been stored. POMCP then draws an underlying state $s$ from this empirical distribution of states that match the current history $H$, and uses $s$ in the simulator \cite{silver2010monte}.

I have found that in our MarketSim environment, occasionally a history is encountered that is uncommon enough that it takes an inordinate length of time for POMCP to generate $minStates$ states that match the history. Often the slowest part of the POMCP algorithm is the rejection sampling procedure for generating states from the empirical distribution associated with a certain history. In some environments, it is much faster to use a domain-specific procedure for generating matching states.

\section{Experiments} \label{experiments}

\subsection{Validation of RL Implementations}

I have performed thorough testing to confirm that the RL methods I have implemented in the EqRegret suite are able to learn at least nearly as efficiently as suggested by prior work. Our tests employ multiple example problems from the RL literature, including MDPs and POMDPs.

To test the classical RL algorithms Q-learning, Sarsa$(\lambda)$, and Q-learning with experience replay, I evaluated their performance on two example problems, as well as in MarketSim.

For a simple MDP problem, I selected Sutton's gridworld, because it has been extensively studied with the Q-learning method \cite{littman1994memoryless} \cite{sutton1990integrated}. In Sutton's gridworld, an agent seeks the shortest path in a maze from a randomly assigned initial state to a goal state. The domain is fairly simple because moving in any direction has the same cost, which never changes. But there are barriers that block the agent from entering certain grid locations, and the agent does not initially know the coordinates of the goal.

For a more challenging MDP problem, I selected the Sailing domain, which has been used several times as a test of MCTS planners \cite{peret2004line} \cite{kocsis2006bandit}. The Sailing MDP is a stochastic shortest path (SSP) problem, where a sailboat agent seeks a shortest path to a goal state under randomly shifting wind directions, which affect the cost of moving in each direction. The Sailing domain is further complicated by a prohibition on sailing directly into the wind, and by a penalty for \emph{tacking} (causing the wind to blow from the opposite side of the boat). I use the Sailing problem to test our POMCP implementation as well as Q-learning and Sarsa$(\lambda)$.

To test POMCP further, I use the board game Battleship, as suggested by Silver and Veness \cite{silver2010monte}. Battleship is a POMDP when played in ``solitaire mode,'' with only one player attempting to sink ships, which are placed according to some random distribution. Randomness comes from the initial distribution of enemy ship locations, and partial observability comes from the unseen locations of the ships on a $10 \textrm{ x } 10$ grid. The player observes only whether a strike at a certain pair of grid coordinates is a \emph{hit} or \emph{miss}. The goal is to minimize the expected number of strikes to sink all opponent ships.

Finally, I tested POMCP with the RockSample[7,8] problem \cite{smith2004heuristic}, which was also used by Silver and Veness to test POMCP \cite{silver2010monte}. This problem has a Mars Rover agent that can move around in a grid world. The agent can use a sensor to obtain noisy readings of whether rocks at different grid locations are ``good,'' meaning that they will produce a positive reward if ``sampled.'' The closer the agent is to a rock, the less noisy the sensor reading is. If the rover moves to a rock's location, it can sample the rock, which yields a positive reward for a ``good'' rock on the first sample, and a negative reward otherwise. The rover gets a large reward for moving to a goal state. A discount factor less than $1$ incentivizes the agent to move to the goal quickly, without taking the time to sense each rock from up close.

\subsection{MarketSim Parameters}

\subsubsection{MarketSim Environments}

I evaluated our learning methods in three MarketSim environments, which differ from each other in the total number of time steps $T$ per simulation and the background trader arrival rate $\lambda_{BG}$. These environments are examples taken directly from previous work on the effect of a market maker on the welfare of other traders \cite{wah2015welfare}. Specifically, I use the A and B environments, with either $1000$ or $4000$ time steps. I call our environments A-1k, B-1k, and A-4k.

Across all three environments, there is one market maker, which arrives according to an exponential interarrival time process with $\lambda_{MM} = 0.005$.

In the A-1k and A-4k environments, each background trader arrives with $\lambda_{BG} = 0.0005$, or one-tenth as frequently on average as the market maker. In the B-1k environment, each background trader arrives with $\lambda_{BG} = 0.005$. The self agent (a background trader) has the same arrival process as the other background traders.

The fundamental value reverts toward the mean according to parameter $\kappa = 0.05$, with jump variance $\sigma_s^2 = 10^6$.

To give an idea of what these parameter settings mean, note that in the A environments, each background trader, including the self agent, is expected to arrive about once per $2000$ time steps. In the A-1k environment, the self agent does not arrive at the market at all in about one third of simulation runs, and the self agent is expected to arrive $0.5$ times. In the A-4k environment, the agent is expected to arrive twice. In the B environments, each background trader is expected to arrive every $200$ time steps, such that the self agent is expected to arrive $5$ times. Therefore, in the A-1k environment, there is likely little value in planning ahead for future arrivals, because the agent is fortunate to arrive even once at the market and unlikely to arrive again. In the A-4k and B-1k environments, an agent could benefit more by considering the likely effects of its subsequent arrivals when deciding how to act.

\subsubsection{Equilibria from EGTA}

Our previous studies of market maker effects in the A-1k, A-4k, and B-1k environments have yielded a few (presumed) role-symmetric Nash equilibria in each environment. These equilibria were discovered with respect to a restricted strategy set of $13$ background trader strategies and up to $7$ market maker strategies. The ZI strategies are listed in Table \ref{tab:zi-strategies}, with the minimum $R_{min}$ and maximum $R_{max}$ of the range of surplus demanded, and the threshold $\eta$ for when to place a market order instead of a limit order. The market maker strategies are shown in Table \ref{tab:mm-strategies}, including the spread $minSpread$ between the innermost rungs of the order ladder, the number of rungs $numRungs$ in the ladder, and the distance between rungs $rungSize$ in the ladder.

\label{sec:nomm}

\begin{table*}[ht]                       
      \centering
      \begin{tabular}{l|ccccccccccccc}
\( R_{\min} \) & 0   & 0   & 0   & 0   & 0   & 0   & 0    & 0    & 0    & 0    & 250 & 500  & 1000 \\
      \hline
\( R_{\max} \) & 65 & 125  & 125 & 250 & 250 & 500 & 1000 & 1000 & 1500 & 2500 & 500 & 1000 & 2000 \\
      \hline
\( \eta \)  & 0.8 & 0.8 & 1   & 0.8 & 1   & 1   & 0.8  & 1    & 0.6  & 1    & 1   & 0.4  & 0.4 \\
      \end{tabular} 
      \caption{ZI strategies included in EGTA equilibration.} 
      \label{tab:zi-strategies}
\end{table*}

\begin{table*}[ht]                       
      \centering
      \begin{tabular}{l|ccccccc}
$numRungs$ & 100 & 100 & 100 & 100 & 100   & 100  & 100 \\
      \hline
$rungSize$ & 25  & 50  & 50  & 50  & 50    & 50   & 100 \\
      \hline
$minSpread$  & 256 & 64  & 128 & 256 & 512   & 1024 & 512 \\
      \end{tabular} 
      \caption{MM strategies included in EGTA equilibration.} 
      \label{tab:mm-strategies}
\end{table*}

From each of the three environments, I have selected two mixed strategies to apply our learning methods to, one which is an equilibrium from EGTA, the other of which is an arbitrary mixed strategy. I will seek to evaluate whether our learning methods can find any strategy that yields higher performance than the equilibrium strategies, refuting the notion that it is an equilibrium in the expanded strategy space of the learner. In contrast, I will use the arbitrary (non-equilibrium) mixed strategies as validity checks that our learning algorithms can successfully discover strategies yielding greater payoff.

\begin{table*}[ht]                       
      \centering
      \begin{tabular}{l|ll|l}
      Mixture & ZI 1 & ZI 2 & Market maker \\
       \hline
      \hline
      A-1k-eq & $0,250,0.8$ & $0,500,1$ & $100,50,512$ \\
       & $0.6868$ & $0.3231$ & $1.0$ \\
      \hline
      A-1k-arb & $0,65,0.8$ & $500,1000,0.4$ & $100,50,512$ \\
       & $0.5$ & $0.5$ & $1.0$ \\
     \hline
      B-1k-eq & $0,500,1$ & $250,500,1$ & $100,100,512$ \\
       & $0.3674$ & $0.6326$ & $1.0$ \\
      \hline
      B-1k-arb & $0,65,0.8$ & $0,250,1$ & $100,100,512$ \\
       & $0.5$ & $0.5$ & $1.0$ \\
      \hline
      A-4k-eq & $0,125,1$ & $0,250,1$ & $100,100,512$ \\
       & $0.1298$ & $0.8702$ & $1.0$ \\
      \hline
      A-4k-arb & $0,125,0.8$ & $500,1000,0.4$ & $100,100,512$ \\
       & $0.5$ & $0.5$ & $1.0$ \\
      \end{tabular} 
      \caption{Mixed strategies for other agents used in RL. Each mixture is either an equilibrium (eq) from EGTA, or an arbitrary mixture (arb). The second and third columns list the background traders' mixed strategy. Each ZI strategy is shown as $R_{min},R_{max},\eta$, with its probability below. Each market maker strategy is shown as $numRungs,rungSize,minSpread$, with its probability below.} 
      \label{tab:mixed-strategies}
\end{table*}

In Table \ref{tab:mixed-strategies}, I present the mixed strategies against which I will apply our learning methods. For each environment, I have selected one equilibrium from EGTA (named with ``eq'' in its title), and I have also constructed an arbitrary mixed strategy, as an equal mixture of two ZI strategies for the background traders, with the same market maker strategy as the chosen equilibrium.

\subsubsection{Action Space for RL}

I selected different sets of actions for the action space $A$ in each RL environment, based on the probability distribution of background trader surplus demands in the equilibrium from EGTA. In the A-1k and A-4k settings, I used an action space of surpluses demanded in $\{30,60,120,240,360\}$. In the B-1k settings, I used an action space of surpluses demanded in $\{20,50,100,200,400\}$. In all cases, $NOOP$ was also an allowed action.

I experimented with two variations of the action space. In the $NoFlip$ setting, the learning agent was allowed to choose whether to place a BUY or SELL order. In the $FlipKnown$ setting, the learning agent was assigned uniformly randomly, independently on each arrival, to buy or sell; the agent chose a different action to select for each of BUY and SELL. The $FlipKnown$ setting is closer to the one faced by ZI traders, which are assigned a role of buyer or seller on each arrival.

\subsubsection{Tile Boundaries for RL}

Through a combination of intuition and trial and error, I decided to use three observation variables as the inputs for our learning agent. First, the learner would observe the surplus the agent would obtain if it bought immediately at the ASK price, based on its marginal private value for the acquired good, the expected final fundamental value $\hat{r}_T$, and the ASK price itself. Second, the learner would observe the surplus it would obtain if it sold immediately at the BID price. Finally, the agent would observe the number of time steps remaining in the simulation, $T - t$.

In each of our three environments, I ran many iterations of the equilibrium mixed strategy profile to obtain summary statistics about the distribution of the $surplusBuyAtAsk$ and $surplusSellAtBid$ statistics. Based on these results, I designed a set of threshold values that would produce an efficient tiling for all three of the environments. This tiling has thresholds for the surplus available along each of the $surplusBuyAtAsk$ and $surplusSellAtBid$, as follows: $(-300, -250, -130, -80, -40, 0, 40, 80)$. These $8$ thresholds divide the real number line into $9$ regions along each dimension, or $81$ buckets in the two dimensions together.

Along the time dimension, I divide the time remaining into $4$ equal-length regions. For the A-1k and B-1k environments, the thresholds for time are $(250,500,750)$, while for the A-4k environment, the time thresholds are $(1000,2000,3000)$. Overall, there are $324$ tiles in the observation space. The tiles are designed such that most of the tiles are visited frequently.

\subsection{Baseline Controls}

I use a few different baseline values for comparison with our learned strategies in each environment. In each case, I will compare the baseline payoff to the expected payoff for one agent playing a learned strategy, while all other agents play the previous mixed strategy from Table \ref{tab:mixed-strategies}.

First, I compare against the expected payoff of playing the other-agent mixed strategy. In the case of EGTA equilibrium strategies like A-1k-eq, I am interested in testing whether the strategy is an equilibrium in the larger strategy space of the learner. If the learner can perform better than the equilibrium strategy, I have refuted the equilibrium in our larger policy space. If, however, no learned strategy outperforms the equilibrium strategy, this lends support to the hypothesis that the strategy was a stable mixed strategy in a more general sense than the restricted strategy space considered by EGTA. In the case of arbitrary strategies like A-1k-arb, the learner should be able to outperform the mean payoff of the strategy.

Next, I compare against the mean payoff of a uniform random agent, which selects actions from the action space of our reinforcement learners. This agent uniformly randomly chooses from among $NOOP$, and buying or selling at any demanded surplus in the set of surpluses to demand for the environment. The uniform random agent plays against other agents playing a mixed strategy from Table \ref{tab:mixed-strategies}.

Finally, I compare against the mean payoff of each pure strategy from the set of ZI strategies in Table \ref{tab:zi-strategies}. Note that these strategies are not in the policy space of the reinforcement learner, so it is conceivable that some ZI strategies from the table are stronger than any learnable policy, although perhaps this is unlikely. It is still useful to compare whether the reinforcement learner performs better than most or all of the pure strategies from the original ZI policy space.

\subsection{RL Methods}

\subsubsection{Q-Learning}

For Q-learning, I use an $\epsilon$-greedy exploration policy, with $\epsilon = 0.2$. In other words, during training the agent takes a uniform-random action from the action set $A$ with probability $0.2$, and plays the greedy action with probability $0.8$.

The current greedy policy is recorded periodically during training, and later the sample mean payoff of each such policy is found over many simulation runs of the greedy policy.

\subsubsection{Sarsa$(\lambda)$}

In Sarsa$(\lambda)$, I also use $\epsilon$-greedy exploration with $\epsilon = 0.2$. I use $\lambda = 0.9$ as the eligibility trace decay factor. That is, each transition in the current playout history receives only $0.9$ as great a $Q$-value update as its successor.

\subsubsection{Tile Coding}

For tile coding, I use either $1$ tiling or $3$ tilings. In the version with 1 tiling, all $324$ observation buckets described above are treated as distinct tiles.

In the version with $3$ tilings, the tiles are arranged as follows. Recall that each point $o$ in observation space $O$ appears in exactly one tile in each of the $3$ tilings.

The first tiling $t_1$ has time thresholds at $(\frac{T}{4},\frac{2T}{4},\frac{3T}{4})$. The second tiling $t_2$ has one time threshold, at $\frac{T}{2}$. The third tiling $t_3$ has no time thresholds, but groups together all states along the time dimension.

The first tiling $t_1$ has surplus thresholds of $(-130, 0)$, along each of the $surplusBuyAtAsk$ and $surplusSellAtBid$ dimensions. The second tiling $t_2$ has surplus thresholds of $(-300, -80, 40)$. The third tiling $t_3$ has surplus thresholds of $(-250, -40, 80)$. Note that the partition of the observation space that results, if I group together each regions that matches the same tile in each tiling, is identical to the partition from the single-tiling framework. This makes results from the single-tiling framework and the three-tiling framework easily comparable.

\subsubsection{Experience Replay}

I experimented with Q-learning both with and without experience replay. When I used experience replay, I played back experiences in batches of $400$, with $100$ new sample runs in between replays. I used a replay memory of $1000$ runs, in a training period of \num[group-separator={,}]{100000} runs total. I selected a small number of replays per batch so that the effect of experience replay could be observed early in the learning curve, within the first $1000$
training runs.

\subsubsection{POMCP}

In the POMCP method, I limited the resources of POMCP by placing a maximum number of playouts per action. I experimented with various numbers of playouts per action, in the set $\{10,100,1000\}$. I set the minimum number of state samples per node at $100$. I initialized the value estimate for each node in the game tree as the expected value of playing the equilibrium profile in Table \ref{tab:mixed-strategies} for that environment, with an initial sample count of $1$. Particle invigoration was not used in the MarketSim environments.

\subsection{Outcome Evaluation}

I evaluated the performance of each learning algorithm in each environment as follows. For POMCP, I ran the POMCP learner with playouts per action of each of $\{10,100,1000\}$, in each environment, over many simulation runs. I recorded all self-agent payoffs in these runs to find the sample mean and $95\%$ bootstrap confidence bounds on the true mean. I then compared these payoffs to the baseline payoffs for the other-agents mixed strategy (playing against itself), the random policy, and each of the $13$ ZI policies from the prior EGTA study.

For Q-learning and Sarsa$(\lambda)$, with or without tile coding or experience replay, I wanted to track the mean performance of learned policies both across simulation runs and across training runs. To that end, I recorded the current greedy policy of the RL learner after each number of training runs in $\{40, 70, 100, 400, 700, 1000, 4000, 7000, 10000, 40000, 70000, 100000\}$. Each of these training durations was sampled from $10$ to $100$ times per environment (not necessarily independently). For each sampled policy after a particular training duration, an equal number of simulation runs were taken of that policy (against other agents playing the mixed strategy against which the policy was learned). The sample mean payoff, along with $95\%$ bootstrap confidence bounds for the mean, were obtained for each training duration. The learning curve could then be compared against the payoffs of the various baseline policies.

\section{Results} \label{results}

\subsection{Validation of RL Methods}

In order to validate that our implementation of the POMCP algorithm performs as reported in prior work, I tested it on three domains: Battleship, RockSample[7,8], and Sailing.

\begin{figure*}
\centering
\includegraphics[width=0.75\columnwidth]{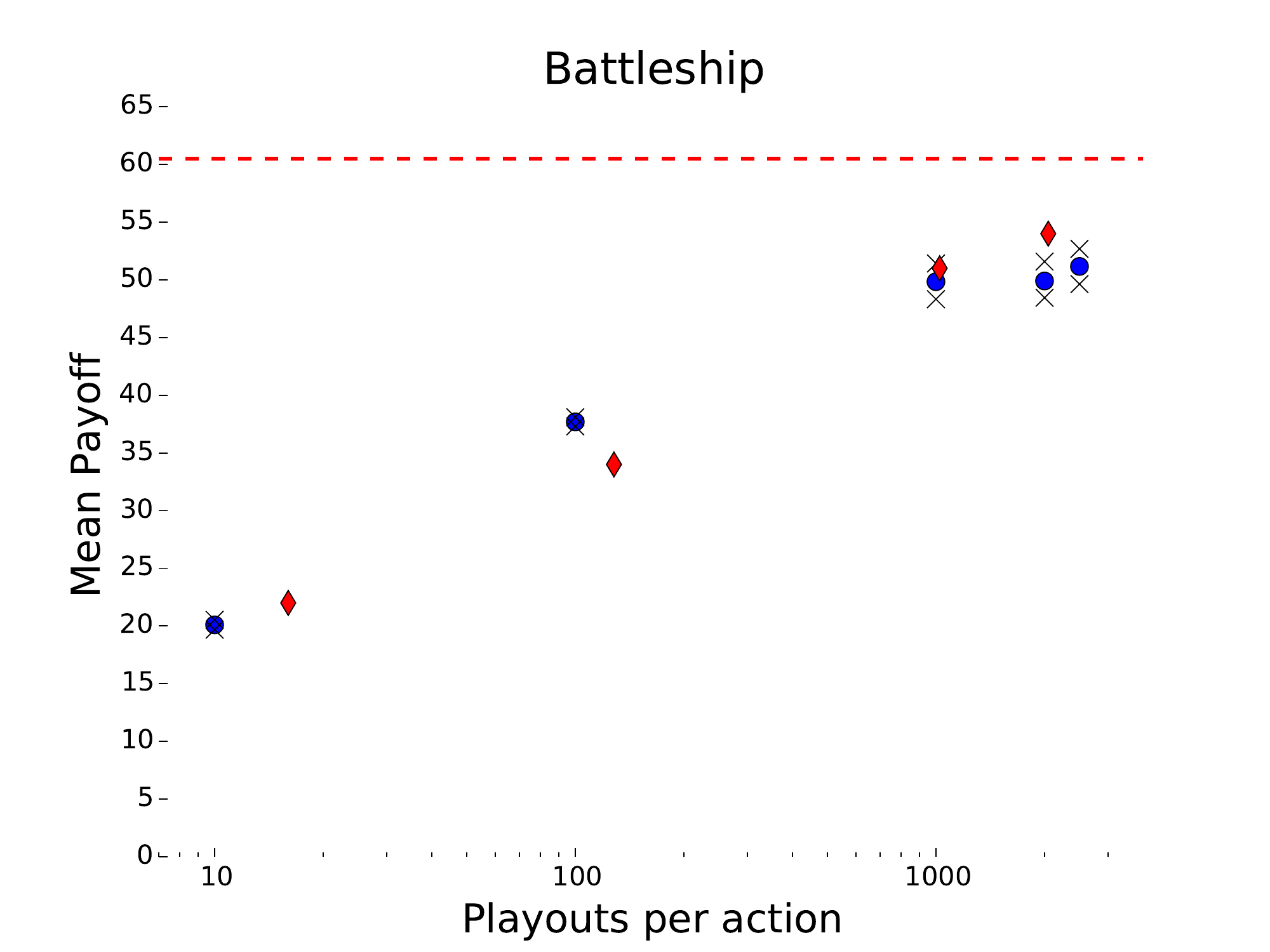}
\caption{Our implementation of POMCP produces roughly the same mean payoff, for a given number of playouts per action, as reported in Silver and Veness. Blue circles show mean payoffs from our implementation. Red diamonds show payoffs from Silver and Veness. Black crosses show $95\%$ confidence intervals for the mean. The red dashed line shows the best performance obtained by Silver and Veness, at $61$.}
\label{fig:battleship}
\end{figure*}

In the Battleship domain, our POMCP implementation achieved a mean payoff (in moves remaining at game completion) of $49.9$ after $2000$ playouts per action, compared with a reported payoff of $51$ after $2048$ moves per action reported by Silver and Veness \cite{silver2010monte}. See Figure \ref{fig:battleship}.

In the RockSample[7,8] domain, our POMCP implementation achieves an average payoff of $13.2$ after $1000$ playouts per action, compared with $14$ after $1024$ playouts per action as reported by Silver and Veness \cite{silver2010monte}. In both the RockSample and Battleship problems, however, the implementation of Silver and Veness achieved higher payoffs than our implementation, when the number of playouts per action was increased beyond $2500$. Silver and Veness achieved greater maximum expected payoffs, possibly through better feature engineering in their rollout policies and node value initialization code.

Results for the Sailing[20,20] domain showed a clear convergence to a stable maximum performance as more POMCP iterations per action were used, but prior work on Sailing did not provide clear explanations of the payoffs achieved for comparison.

\begin{figure*}
\centering
\includegraphics[width=1.0\columnwidth]{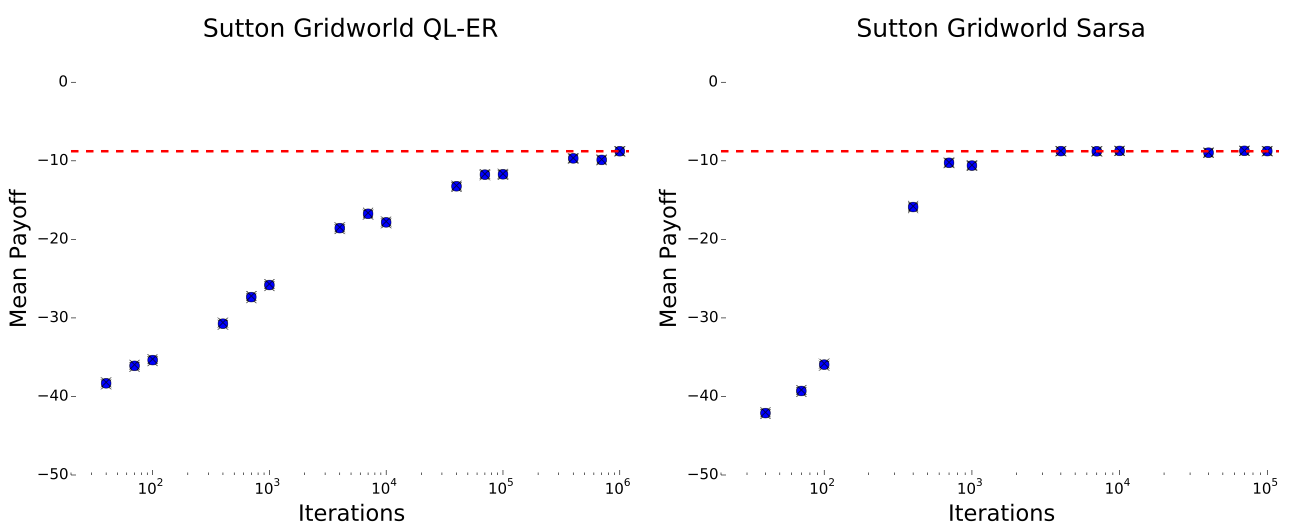}
\caption{Our implementations of Q-learning with experience replay and Sarsa$(\lambda)$ converge to the optimal payoff in Sutton's gridworld, but Sarsa$(\lambda)$ converges much faster. The dashed red line shows the optimal payoff, $-8.78$. Blue circles show the mean payoff for our implementation. Black crosses show $95\%$ confidence intervals.}
\label{fig:suttonGridworld}
\end{figure*}

I tested our implementations of Q-learning and Sarsa$(\lambda)$ in two domains, the Sailing domain and Sutton's gridworld. In Sutton's gridworld, there is a known optimal payoff in the mean over all initial states, of $-8.78$, meaning it takes $8.78$ steps on average to reach the goal \cite{littman1994memoryless}. Our Q-learner reached its best payoff on average at $-9.35$ after \num[group-separator={,}]{700000} simulation runs (see Figure \ref{fig:suttonGridworld}). The Q-learner with experience replay performed better, reaching an optimal performance of $-8.78$ after \num[group-separator={,}]{1000000} simulation runs. Best of all was our Sarsa$(\lambda)$ implementation, which reached an optimal performance if $-8.78$ on average after only $4000$ simulation runs.

Our classical RL algorithms, Q-learning and Sarsa$(\lambda)$, did not perform nearly as effectively as POMCP on the Sailing domain stochastic shortest path problem. These methods' performance plateaued at approximately $-500$ as the mean payoff, while POMCP plateaued at $-68$. This is not an indication of implementation issues with our algorithms, so much as of the known difficulty of classical RL algorithms with SSP problems.

\subsection{Q-Learning and Sarsa$(\lambda)$}

\subsubsection{NoFlip Setting}

In the NoFlip setting, where the self agent is allowed to choose whether to submit a buy or sell order on each arrival, all the classical learning agents tested easily outperformed the EGTA equilibrium payoffs. See Table \ref{tab:nf-classical-payoffs}.

\begin{table*}[ht]                       
      \centering
      \begin{tabular}{l|l|ll}
      Mixture & Equilibrium & Sarsa$(\lambda)$ 3-tiling & Q-Learning 3-tiling \\
       \hline
      \hline
      A-1k-eq & 63.74 & 116.30 & 117.90 \\
     \hline
      B-1k-eq & 426.65 & 561.23 & n.a. \\
      \hline
      A-4k-eq & 231.88 & 385.13 & n.a. \\
      \end{tabular} 
      \caption{For every EGTA equilibrium tested, the tested classical RL methods produced dramatically higher mean payoff, in the NoFlip setting where RL agents are allowed to choose whether to buy or sell. The second column shows the mean payoff of each mixed strategy found by EGTA. The right columns show the best mean payoff found during \num[group-separator={,}]{100000} training iterations of the stated learning method. Means are taken over at least $10$ training cycles.
}
      \label{tab:nf-classical-payoffs}
\end{table*}

\begin{figure*}
\centering
\includegraphics[width=0.7\columnwidth]{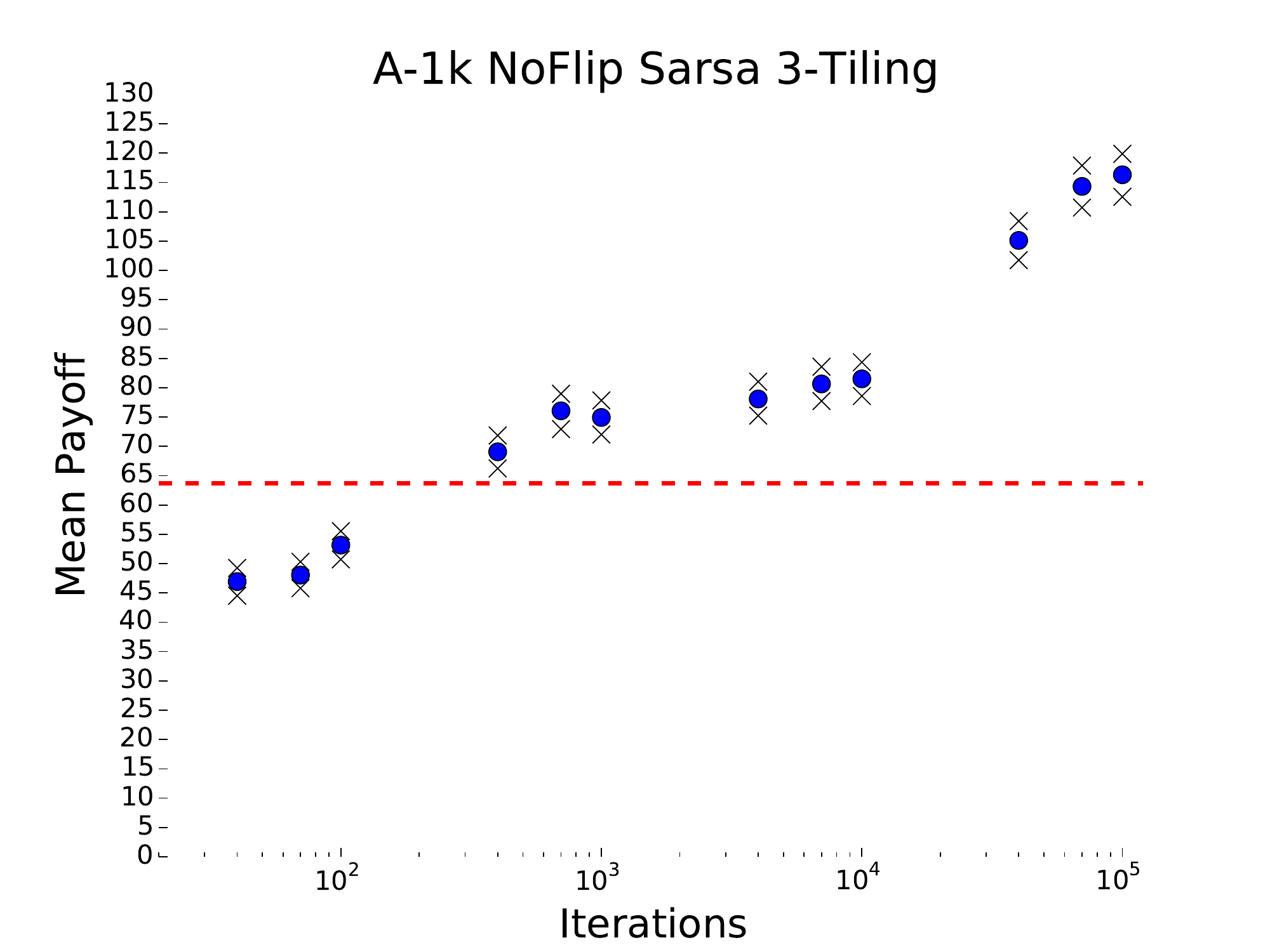}
\caption{In the A-1k NoFlip environment, Sarsa$(\lambda)$ easily outperforms the equilibrium payoff from EGTA. The dashed red line shows the mean payoff of the EGTA policy. The blue circles show sample mean payoffs for learned policies at various stages of training. The black crosses show $95\%$ confidence intervals.}
\label{fig:a1_NF_SLTile}
\end{figure*}

Not only do classical RL algorithms produce better mean payoffs than the EGTA payoff in NoFlip setting, the algorithms produce steady learning curves that tend to be almost strictly increasing until a plateau. See Figure \ref{fig:a1_NF_SLTile} for an example learning curve, for Sarsa$(\lambda)$ with $3$ tilings, in the A-1k NoFlip environment. The learner statistically significantly outperforms the opponents' mixed strategy within $400$ training steps, and eventually almost doubles the opponents' expected payoff.

I should note that in the NoFlip setting, the learning agent is given an important advantage over other agents, as the learner is allowed to choose whether to place BUY or SELL order at each arrival, where the other background traders are forced to do one or the other, based on a fair ``coin flip.'' In the A-1k environment, each background trader is likely to arrive at the market only once, and at that time, it may be possible to make an advantageous trade immediately to buy or to sell, but not both. As a result, it is natural that our learning agents achieve roughly double the mean payoff of the equilibrium mixed strategy of A-1k-eq, or $117.90$ instead of $63.74$. In the A-4k and B-1k environments, each background trader arrives $2$ or $5$ times in expectation, so it is not as great an advantage to be allowed to choose buy or sell at each arrival. Still, the learning agent in the NoFlip case achieves much higher payoffs than the A-4k-eq or B-1k-eq background traders.

\subsubsection{FlipKnown Setting, Equilibrium Policy}

The FlipKnown setting is inherently more challenging, as the learning agent is not allowed to choose whether to place a BUY or SELL order at each arrival at the market. Instead, the agent decides how much surplus to demand as buyer and how much as seller, and then a coin flip decides whether the agent will submit its selected BUY or SELL order. Although this environment is more difficult for the learner, it also places the learner on a more even footing with the other background traders, which also have to place BUY or SELL orders based on a coin flip result.

\begin{table*}[ht]                       
      \centering
      \begin{tabular}{l|l|l|ll}
      Mixture & Equilibrium & Random & Sarsa$(\lambda)$ 3-tiling & Q-Learning 3-tiling \\
       \hline
      \hline
      A-1k-eq & 63.74 & 56.27 & 64.59 $\pm$ 2.6 & 65.55 $\pm $ 2.6 \\
     \hline
      B-1k-eq & 426.65 & 381.70 & 422.05 $\pm $ 20 & 422.70 $\pm$ 21 \\
      \hline
      A-4k-eq & 231.88 & 217.52 & 239.94 $\pm$ 12 & 242.24 $\pm$ 13 \\
      \end{tabular} 
      \caption{The best classically learned policies for FlipKnown environments were nearly the same in mean payoff as the EGTA equilibrium policies, and in no case better or worse beyond a $95\%$ confidence interval. Each learner value shown is for the best intermediate learned policy in sample mean payoff, over the $12$ training lengths $(40,70,100,\ldots,40000,70000,100000)$. Each mean payoff is listed along with a bootstrap $95\%$ confidence interval for the mean. The Random column shows the mean payoff of a uniform random policy, over the learner's allowed action set.
}
      \label{tab:fk-classical-payoffs}
\end{table*}

As shown in Table \ref{tab:fk-classical-payoffs}, in every case the classical learning agent obtained a payoff that is not statistically significantly different from the payoff of the EGTA equilibrium. In the A-1k-eq and A-4k-eq environments, both Sarsa$(\lambda)$ with 3 tilings and Q-learning with 3 tilings found some policies that yielded better sample mean payoffs than the original equilibrium, but it is not yet clear whether these policies are genuinely successful deviations from the equilibrium, or only products of selection bias. Each time a training algorithm such as Q-learning is run against the A-1k-eq environment, it generally produces some intermediate policies that appear better than the A-1k-eq policy in an initial sample, but if more payoff samples of these high-performing policies are taken, it turns out they were actually no better, but simply products of multiple testing bias. In our experiments over many independent training runs of Sarsa$(\lambda)$ and Q-learning in the FlipKnown setting, I have not produced any policies that outperform the EGTA equilibrium beyond a $95\%$ confidence bound.

\begin{figure*}
\centering
\includegraphics[width=0.7\columnwidth]{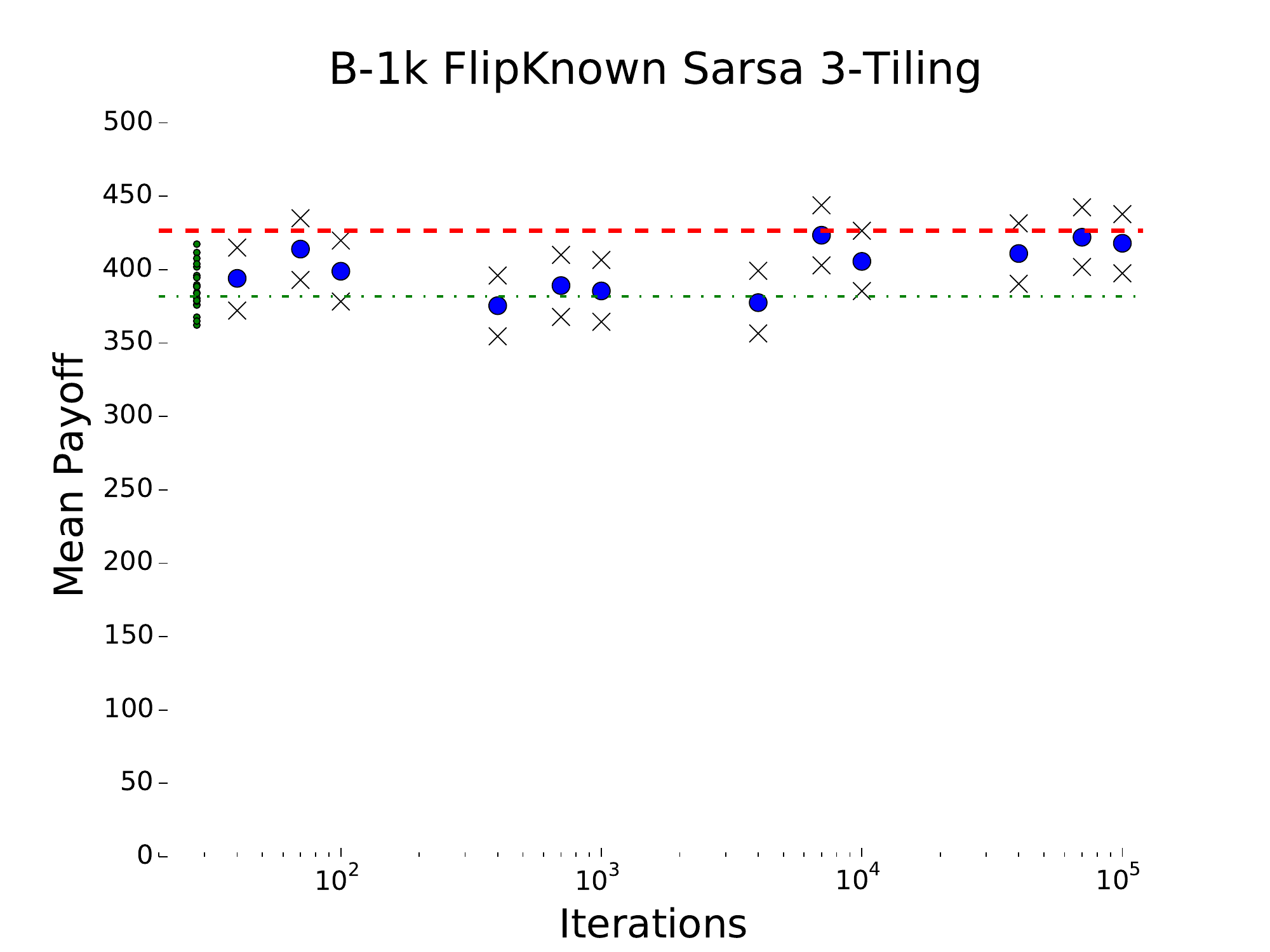}
\caption{In the B-1k FlipKnown environment, Sarsa$(\lambda)$ 3-tiling does not perform stably, but yields mean payoffs that sometimes decrease with increasing training steps. After \num[group-separator={,}]{100000} training steps, mean payoff is not significantly different from the payoff of the EGTA background trader policy. Dashed red line shows EGTA background trader payoff. Dotted green line shows random agent mean payoff. Blue circles show mean learner payoff. Black crosses show $95\%$ confidence intervals. Small green circles show payoffs of $10$ randomly-generated policies.}
\label{fig:b1_FK_SLTileBaseline}
\end{figure*}

\begin{figure*}
\centering
\includegraphics[width=0.7\columnwidth]{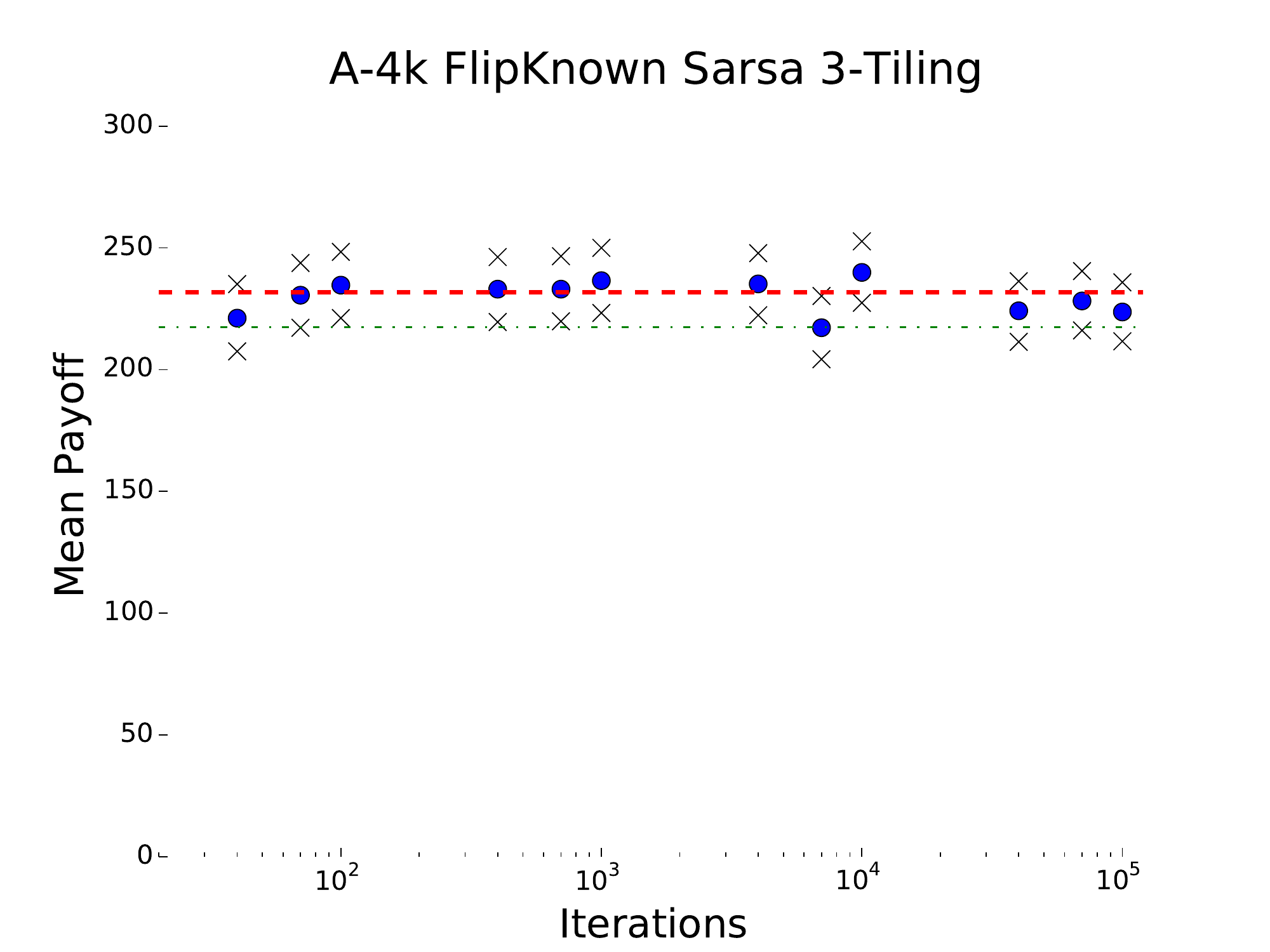}
\caption{In the A-4k FlipKnown environment, Sarsa$(\lambda)$ 3-tiling gives unstable performance, with payoffs that do not monotonically increase with training. After \num[group-separator={,}]{100000} training steps, mean payoff is not significantly different from the payoff of the EGTA background trader policy. Dashed red line shows EGTA background trader payoff. Dotted green line shows random agent mean payoff. Blue circles show mean learner payoff. Black crosses show $95\%$ confidence intervals.}
\label{fig:a4_FK_SLTileBaseline}
\end{figure*}

Unlike in the test environments and the NoFlip setting, our learning algorithms did not yield stable performance in the FlipKnown environment. In a typical learning session, a learner in a FlipKnown equilibrium setting (A-1k-eq, B-1k-eq, or A-4k-eq) would quickly achieve better performance than a randomly-generated agent, within $100$ training runs. Afterward, the agent would erratically produce intermediate greedy policies with better or worse performance, for the first \num[group-separator={,}]{10000} cycles or so. By cycle \num[group-separator={,}]{100000}, performance would often stabilize, though at a level that is not statistically significantly different from the payoff of the EGTA equilibrium strategy. This pattern suggests that the learning algorithms quickly reached a plateau of performance near the performance of the equilibrium policy, and were not able to do any better. This pattern is exemplified in the performance of Sarsa$(\lambda)$ in the B-1k-eq environment, shown in Figure \ref{fig:b1_FK_SLTileBaseline}, or in the A-4k-eq environment, shown in Figure \ref{fig:a4_FK_SLTileBaseline}.

\subsubsection{FlipKnown Setting, Arbitrary Policy}

We have seen that our learning algorithms were not able to discover significantly better policies than the EGTA equilibrium policy, in FlipKnown settings. After this finding, I ran the learning algorithms against arbitrary, non-equilibrium other-agent policies in the FlipKnown setting, for the A-1k, B-1k, and A-4k environments. The policies for the other agents are shown in Table \ref{tab:mixed-strategies}, as A-1k-arb, B-1k-arb, and A-4k-arb. These tests were intended as a sanity check for our algorithms, to ensure that they could discover successful strategy deviations to opponent strategies that do not constitute a Nash equilibrium.

In Table \ref{tab:fk-arb-classical-payoffs}, I show the payoff of the best Q-learning policy against the arbitrary other-agent strategy profiles. It is clear that Q-learning with 3 tilings performs statistically significantly better than both the arbitrary opponent strategy, and a random response to that strategy, in all cases. Note that the reported payoffs for Q-learning are based on the mean over at least $10$ and up to $100$ independent training cycles. 

\begin{table*}[ht]                       
      \centering
      \begin{tabular}{l|l|l|l}
      Mixture & Equilibrium & Random & Q-Learning 3-tiling \\
       \hline
      \hline
      A-1k-arb & 54.68 & 54.25 & 61.11 $\pm$ 2.7 \\
     \hline
      B-1k-arb & 431.71 & 419.63 & 471.85 $\pm $ 21 \\
      \hline
      A-4k-arb & 193.53 & 198.16 & 231.88 $\pm$ 13 \\
      \end{tabular} 
      \caption{The best classically learned policies for FlipKnown environments were statistically significantly better than the arbitrary other-agent policies, as well as better than a random policy playing against those other-agent policies. Each learner value shown is for the best intermediate learned policy in sample mean payoff, over the $12$ training lengths $(40,70,100,\ldots,40000,70000,100000)$. Each mean payoff is listed along with a bootstrap $95\%$ confidence interval for the mean. The Random column shows the mean payoff of a uniform random policy, over the learner's allowed action set. Small green circles show payoffs of $10$ randomly-generated policies.
}
      \label{tab:fk-arb-classical-payoffs}
\end{table*}

\begin{figure*}
\centering
\includegraphics[width=0.7\columnwidth]{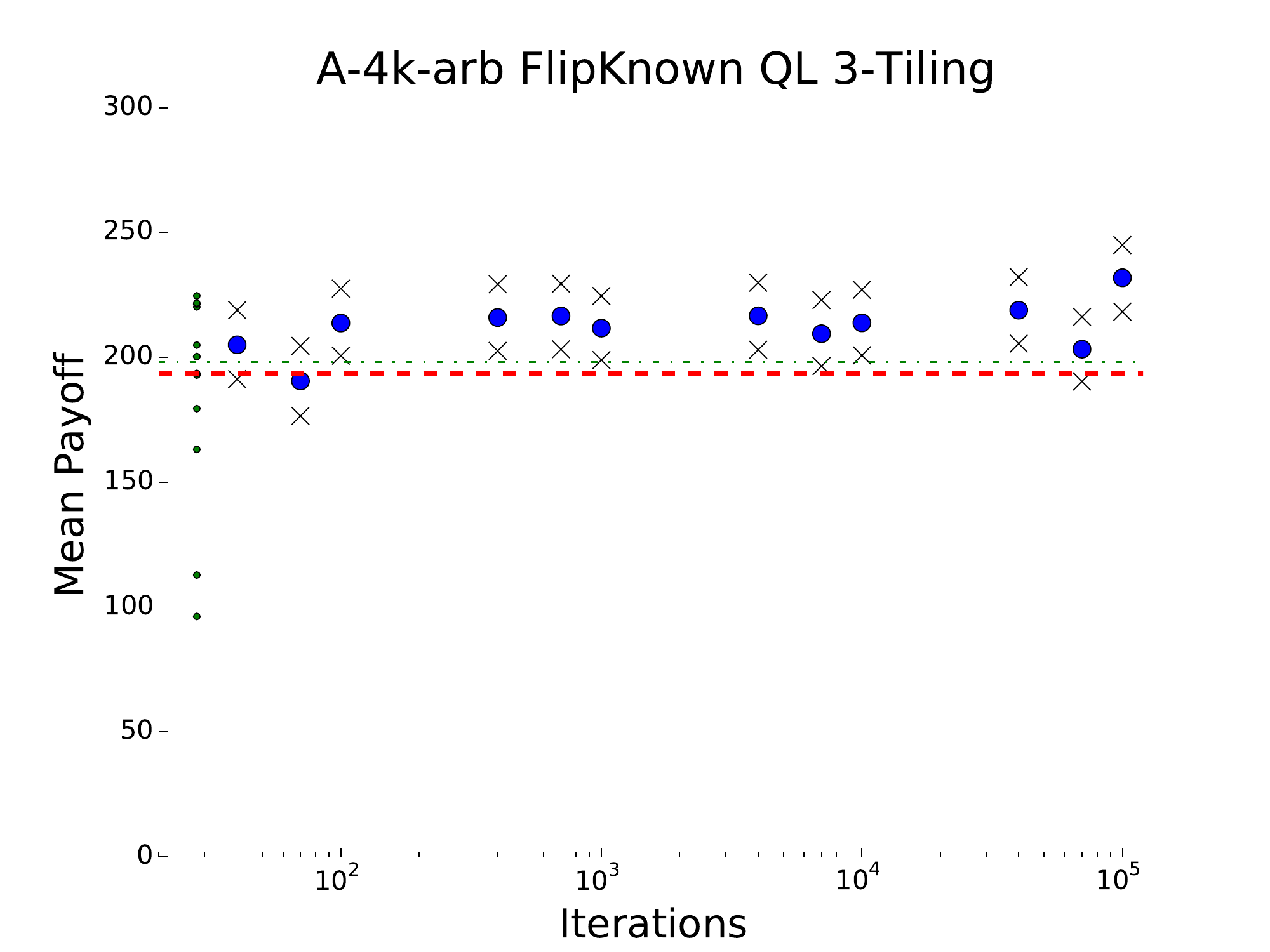}
\caption{In the A-4k FlipKnown environment with the non-equilibrium other-agent policy A-4k-arb, Q-learning 3-tiling produces payoffs that are statistically significantly better than the original policy or a random response. After only $1000$ training steps, mean payoff is significantly better than the payoff of the ZI background trader policy. Dashed red line shows EGTA background trader payoff. Dotted green line shows random agent mean payoff. Blue circles show mean learner payoff. Black crosses show $95\%$ confidence intervals. Small green circles show payoffs of the $13$ ZI pure strategies from EGTA.}
\label{fig:a4_arb_FK_QLTileBaseline}
\end{figure*}

Compared to the equilibrium settings, the payoff of learned policies tended to be more stably increasing with longer training, so there is less risk of exaggerated performance values due to multiple testing bias over the intermediate policies' results. In Figure \ref{fig:a4_arb_FK_QLTileBaseline}, I show the learning curve for the A-4k-arb environment. Note how the Q-learner surpasses the payoff of the baseline strategy within $1000$ training runs, and attains a large margin of about $38$ over the baseline after \num[group-separator={,}]{100000} training runs. Note the small green circles to the left of the figure, indicating the mean payoff of the $13$ ZI strategies from EGTA shown in Table \ref{tab:zi-strategies}. Our learned strategies eventually perform comparably to the best of these pure strategies, which are not within the strategy space of the Q-learner and could conceivably be better than any expressible policy. Even so, after \num[group-separator={,}]{100000} training runs the Q-learner has a policy with greater sample mean payoff than any of the original ZI pure strategies, in this A-4k-arb case.

Our results in the arbitrary other-agent policy settings, A-1k-arb, A-4k-arb, and B-1k-arb, indicate that our classical RL implementations are able to learn superior policies in cases where such a policy is believed to exist in their strategy space. In all three cases, the Q-learner with 3 tilings performed statistically significantly better than the other-agent policy and a random response.

\subsection{POMCP}

POMCP generally gave less efficient performance than the classical RL methods Q-learning and Sarsa$(\lambda)$ in the MarketSim environments. The slowness of the market simulator prevented us from using more than $100$ or $1000$ playouts per action, or from obtaining enough payoff samples in some environments to obtain accurate estimates of the value of the POMCP policy.

\subsubsection{NoFlip Setting}

In the NoFlip setting, POMCP was able to learn a policy with better mean performance than the EGTA equilibrium policies, but statistically significantly so only in the A-1k-eq environment, not the B-1k-eq or A-4k-eq environments. As shown in Table \ref{tab:nf-pomcp-payoffs}, the best POMCP sample mean payoff after $10$, $100$, or $1000$ steps per action was similar to the payoff of the EGTA equilibrium policy in the B-1k-eq and A-4k-eq settings. Moreover, the confidence intervals are extremely wide in those cases, due to the variability of the POMCP method's implicit policy, combined with the high cost of obtaining more sample payoffs from POMCP.

\begin{table*}[ht]                       
      \centering
      \begin{tabular}{l|l|l}
      Mixture & Equilibrium & POMCP \\
       \hline
      \hline
      A-1k-eq & 63.74 & 79.80 $\pm$ 16\\
     \hline
      B-1k-eq & 426.65 & 446.19 $\pm$ 66 \\
      \hline
      A-4k-eq & 231.88 & 235.07 $\pm$ 50 \\
      \end{tabular} 
      \caption{For every EGTA equilibrium tested, POMCP yielded a higher mean payoff, in the NoFlip setting where the POMCP agent is allowed to choose whether to buy or sell. The payoff benefit was statistically significant only in the A-1k-eq setting. The second column shows the mean payoff of each mixed strategy found by EGTA. The right column shows the best mean payoff found by POMCP, with either $10$, $100$, or $1000$ playouts per action. Mean payoffs from POMCP are shown with $95\%$ confidence intervals.
}
      \label{tab:nf-pomcp-payoffs}
\end{table*}

\begin{figure*}
\centering
\includegraphics[width=0.7\columnwidth]{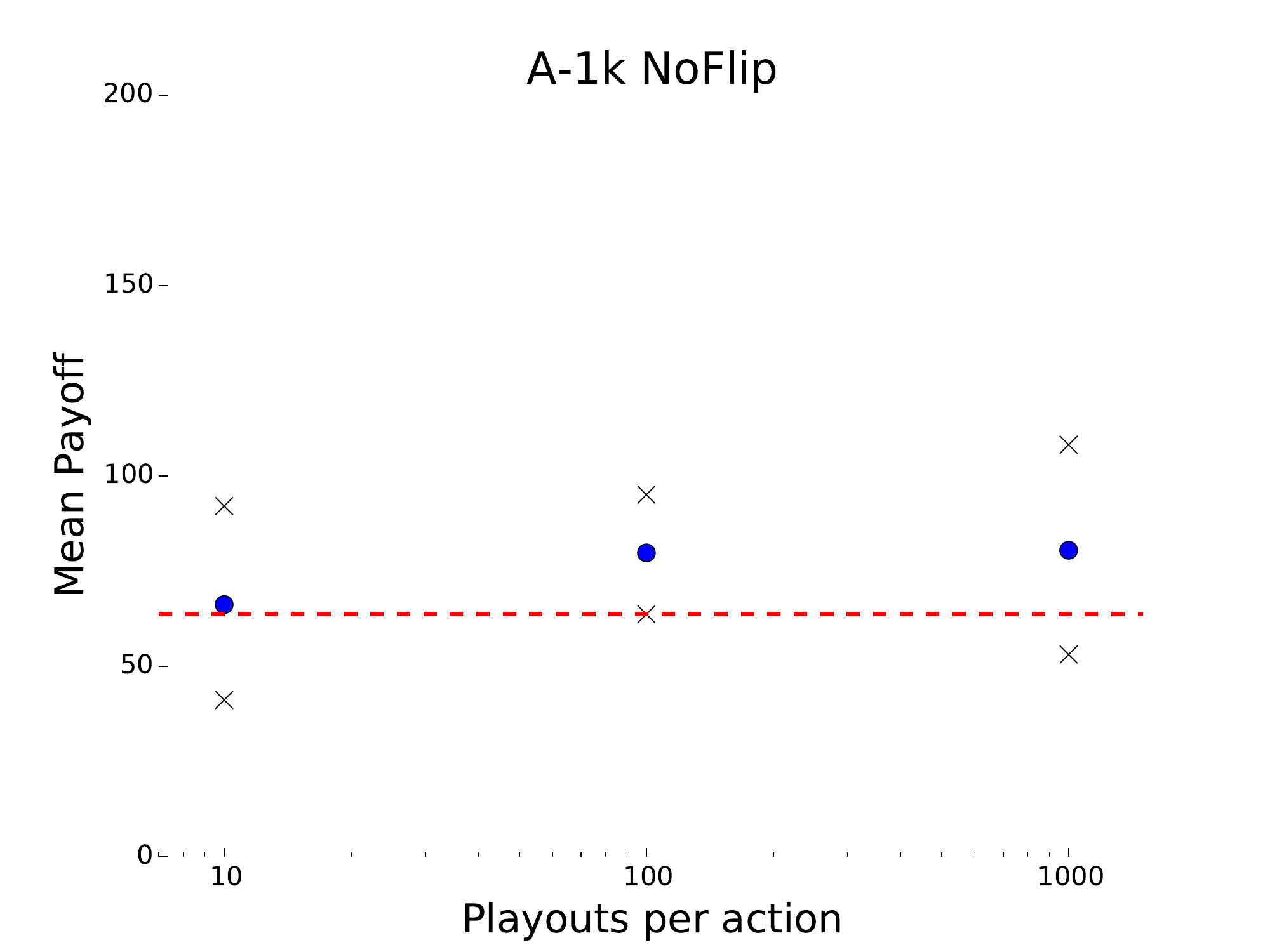}
\caption{In the A-1k-eq NoFlip environment, POMCP produces payoffs that are statistically significantly better than the original policy, and produces better mean payoffs the more playouts are taken per action. The difference is statistically significant for $100$ playouts per action. The confidence interval is wider for $1000$ playouts per action because fewer samples were taken, due to the high cost of running POMCP with such a deep search tree. Dashed red line shows EGTA background trader payoff. Blue circles show mean learner payoff. Black crosses show $95\%$ confidence intervals.}
\label{fig:a1_pomcp_NF}
\end{figure*}

In Figure \ref{fig:a1_pomcp_NF}, I show the performance of POMCP's implicit policy for $10$, $100$, or $1000$ simulations per action. Note how the mean payoff increases with increasing training in this case. For $100$ simulations per action, I obtained more sample payoffs than for the other cases, to demonstrate statistically significantly better performance than the equilibrium policy. (It would have been much more computationally expensive to obtain so many samples for $1000$ playouts per action.)

\subsubsection{FlipKnown Setting, Equilibrium Policy}

\begin{table*}[ht]                       
      \centering
      \begin{tabular}{l|l|l}
      Mixture & Equilibrium & POMCP \\
       \hline
      \hline
      A-1k-eq & 63.74 & 53.52 $\pm$ 25 \\
     \hline
      B-1k-eq & 426.65 & 245.20 $\pm$ 134 \\
      \hline
      A-4k-eq & 231.88 & 236.258 $\pm$ 178 \\
      \end{tabular} 
      \caption{Mean payoffs for POMCP were not better than the EGTA equilibrium payoff in any of the environments tested, and confidence intervals were wide after our sample budget was reached, due to the high cost of running the POMCP policies. I show the POMCP payoff for the number of simulations per action yielding highest mean reward, of $10$, $100$, $1000$. I also show $95\%$ confidence intervals for POMCP payoffs.
}
      \label{tab:fk-pomcp-payoffs}
\end{table*}

In the FlipKnown settings, POMCP performed much less efficiently than Q-learning. Because POMCP does not produce an explicit policy which can be simulated quickly after training, it was very slow and expensive to acquire sample results from POMCP, especially with large numbers of playouts per action. As a result, the confidence intervals for the payoff of POMCP policies are very wide, particularly with high playouts per action.

As shown in Table \ref{tab:fk-pomcp-payoffs}, POMCP did not produce statistically significantly better payoffs than the EGTA equilibrium in any of the environments. In the B-1k-eq environment, POMCP performed significantly worse than the EGTA equilibrium, and worse even than a random baseline policy, for reasons that are not yet known.

\section{Discussion} \label{discussion}

In this work I have applied reinforcement learning tools to evaluate the strategic stability of profiles returned by EGTA. I developed a suite of RL methods, including implementations of POMCP, Q-learning, Sarsa$(\lambda)$, and variants of those techniques with experience replay and tile coding. Experiments verified the correctness of the implementations on example POMDP and MDP problems from prior literature, such as RockSample, Battleship, Sailing, and Sutton's gridworld. The central experiments of this work used RL to challenge equilibrium strategy profiles from EGTA, that were presented in the recent work or Wah and Wellman \cite{wah2015welfare}.

I showed that all of our RL methods reached a plateau of mean payoff at or near the level of the equilibrium profiles from EGTA, in all market environments tested (in the FlipKnown setting that closely mirrors prior work). In general, the classical algorithms Q-learning and Sarsa gave better performance than POMCP, due in part to the high cost in computation time required to run POMCP to an adequate search depth. When I applied the same classical RL methods to non-equilibrium other-agent strategy profiles, the RL methods converged to better-performing policy responses in all tests. Similarly, when I applied RL methods to the EGTA equilibrium profiles, but allowed the learning agent to choose whether to buy or sell at each arrival, the learner was always able to achieve higher mean payoffs than the EGTA policy. This suggests that the learning methods used were effective at searching the policy space for effective strategies, but the equilibrium policies found by EGTA had low regret, as I wanted to show.

There are inherent limitations with our empirical approach to validating a supposed Nash equilibrium profile. It is possible that a reinforcement learning approach will fail to find a beneficial deviating strategy for a profile, even though the profile has high regret, and thus would not be adopted by a rational agent. Therefore, our work is not a logical proof that the strategy profiles from the market maker EGTA study are Nash equilibria, even with respect to the learning agents' strategy space. I have taken several steps, however, to add rigor to our method and support the claim that the EGTA profiles probably have low regret, with respect to an enlarged strategy space. I have used a suite of RL methods that are very different in their mechanism of action; I have validated each of those RL methods in example problems from the literature; and I have shown that those methods find beneficial strategy deviations in our market simulation, under relaxed conditions (choosing to buy or sell, or competing against a non-equilibrium profile). These tests add rigor to our approach for validating the strategic stability of a profile returned by EGTA, and while no such empirical method can conclusively prove a profile is a Nash equilibrium in an infinite strategy space, our method can produce strong supporting evidence of profile stability.

\bibliographystyle{ACM-Reference-Format-Journals}
\bibliography{rlValidate-bibfile}

\end{document}